\documentclass[runningheads]{llncs}

\usepackage{multirow}
\usepackage{amsmath}
\usepackage{makecell}

\newcommand{\pose}[0]{\boldsymbol{\theta}}
\newcommand{\shape}[0]{\boldsymbol{\beta}}
\newcommand{\trans}[0]{\boldsymbol{t}}

\newcommand{\set}[1]{\mathcal{#1}}

\newcommand{\real}[1]{\mathbb{R}^{#1}}

\newcommand{\modelName}{GRAFT}
\newcommand{\scene}[0]{\set{P}} %
\newcommand{\hparams}[0]{\mathbf{\Theta}}
\newcommand{\dhparams}[0]{\Delta\mathbf{\Theta}}

\newcommand{\boldparagraph}[1]{\vspace{0.1cm}\noindent{\bf #1}}

\usepackage[table]{xcolor}
\usepackage{xcolor}
\usepackage{siunitx}
\usepackage{algorithm}
\usepackage[noend]{algpseudocode}
\usepackage{pifont}        %
\usepackage{enumitem}
\usepackage[normalem]{ulem}
\usepackage{wrapfig}

\definecolor{heatYellow}{HTML}{FFFFB2}   %
\definecolor{heatOrange}{HTML}{FFD9B2}   %
\definecolor{heatRed}{HTML}{FFB2B2}   %
\definecolor{Bred}{HTML}{b5341e}   
\definecolor{Fgreen}{HTML}{309c59}

\newcommand{\cmark}{\textcolor{Fgreen}{\ding{52}}} %
\newcommand{\xmark}{\textcolor{Bred}{\ding{56}}}

\definecolor{myPink}{HTML}{E91E63}  %
\definecolor{black}{HTML}{000000}  %
\definecolor{BetaColor}{rgb}{0.8,0,0.8}

\usepackage{eccv}

\usepackage{eccvabbrv}

\usepackage{graphicx}
\usepackage{booktabs}

\usepackage[accsupp]{axessibility}  %

\usepackage[breaklinks,colorlinks,citecolor=eccvblue]{hyperref}

\usepackage{orcidlink}

\begin{document}

\newif\ifincludesupplementary
\includesupplementarytrue

\ifincludesupplementary
  \newcommand{\cameraonly}[1]{}%
\else
  \newcommand{\cameraonly}[1]{#1}%
\fi

\title{GRAFT: Geometric Refinement and Fitting Transformer for Human Scene Reconstruction} 

\titlerunning{GRAFT: Geometric Refinement and Fitting Transformer}

\author{Pradyumna YM\textsuperscript{1,2} \and
Yuxuan Xue\textsuperscript{1,2}\thanks{Corresponding Author} \and
Yue Chen\textsuperscript{3} \and
Nikita Kister\textsuperscript{1,2} \\
Istv{\'a}n S{\'a}r{\'a}ndi\textsuperscript{1,2} \and
Gerard Pons-Moll\textsuperscript{1,2,4}}

\authorrunning{YM et al.}

\institute{\textsuperscript{1}University of T\"ubingen\cameraonly{, Germany} \quad
\textsuperscript{2}T\"ubingen AI Center\cameraonly{, Germany} \\
\textsuperscript{3}Westlake University\cameraonly{, China} \quad
\textsuperscript{4}Max Planck Institute for Informatics\cameraonly{, Germany}}

\makeatletter
\renewcommand{\@fnsymbol}[1]{\ensuremath{\ifcase#1\or\dagger\or *\or\ddagger\or\mathsection\or\mathparagraph\or\|\or **\or\dagger\dagger\or\ddagger\ddagger\else\@ctrerr\fi}}
\makeatother
\renewcommand{\thefootnote}{\fnsymbol{footnote}}
\maketitle
\setcounter{footnote}{0}
\renewcommand{\thefootnote}{\arabic{footnote}}

\begin{abstract}
  Reconstructing physically plausible 3D human-scene interactions (HSI) from a single image currently presents a trade-off: optimization based methods offer accurate contact but are slow (${\sim}$20\,s), while feed-forward approaches are fast yet lack explicit interaction reasoning, producing floating and interpenetration artifacts.
  Our key insight is that geometry-based human--scene fitting can be amortized into fast feed-forward inference. We present \textbf{GRAFT} (\textbf{G}eometric \textbf{R}e\-fine\-ment \textbf{A}nd \textbf{F}it\-ting \textbf{T}rans\-form\-er), a learned HSI prior that predicts \emph{Interaction Gradients}: corrective parameter updates that iteratively refine human meshes by reasoning about their 3D relationship to the surrounding scene.
  GRAFT encodes the interaction state into compact body-anchored tokens, each grounded in the scene geometry via \emph{Geometric Probes} that capture spatial relationships with nearby surfaces.
  A lightweight transformer recurrently updates human meshes and re-probes the scene, ensuring the final pose aligns with both learned priors and observed geometry.
  GRAFT operates either as an end-to-end reconstructor using image features, or with geometry alone as a transferable plug-and-play HSI prior that improves feed-forward methods without retraining. 
  Experiments show GRAFT improves interaction quality by up to 122\% over state-of-the-art feed-forward methods and matches optimization-based interaction quality at ${\sim}100{\times}$ lower runtime, while generalizing seamlessly to in-the-wild multi-person scenes and being preferred in 64.8\% of three-way user study. Project page: \url{https://pradyumnaym.github.io/graft}.


\end{abstract}

\section{Introduction}

\label{sec:intro}
Holistic 3D understanding of humans and their surrounding environments is essential for emerging technologies such as embodied AI, sports analytics, and augmented reality. 
Such applications require more than just the correct spatial localization of subjects; they demand explicit \emph{interaction reasoning}—the ability to understand how a human relates to the 3D physical constraints of a scene to reconstruct coherent contact and reason about affordances. While monocular human mesh recovery (HMR) has seen significant advances, estimating physically plausible human-scene alignment remains a formidable challenge: depth-scale ambiguity compounds with incompatible learned biases across separate human and scene models, so their outputs are rarely metrically or geometrically consistent with one another.

Current state-of-the-art HSI reconstruction methods~\cite{ym2025physic,liu2025joint,mueller2024hsfm} tackle these challenges using off-the-shelf human-scene initialization, and then optimizing them against energy functions that encourage realistic HSI.
However, this optimization is expensive (${\sim}$20\,s per instance~\cite{ym2025physic}), susceptible to local minima due to its reliance on analytical gradients, and unable to learn generalizable interaction priors, as geometry enters only as a hard test-time constraint.

To improve inference speed, recent feed-forward methods~\cite{chen2025human3r, li2026unish} build on foundational pose and depth models to jointly regress humans and scenes in a single pass.
However, they perform no explicit interaction reasoning: scene geometry is never queried during pose decoding, so the model relies entirely on learned priors—producing meshes that frequently float above or penetrate the very scene it reconstructed.
In practice, a costly test-time optimization such as PhySIC~\cite{ym2025physic} is still required to obtain accurate HSI reconstructions.

\begin{figure}[t]
    \centering
    \includegraphics[width=1\linewidth]{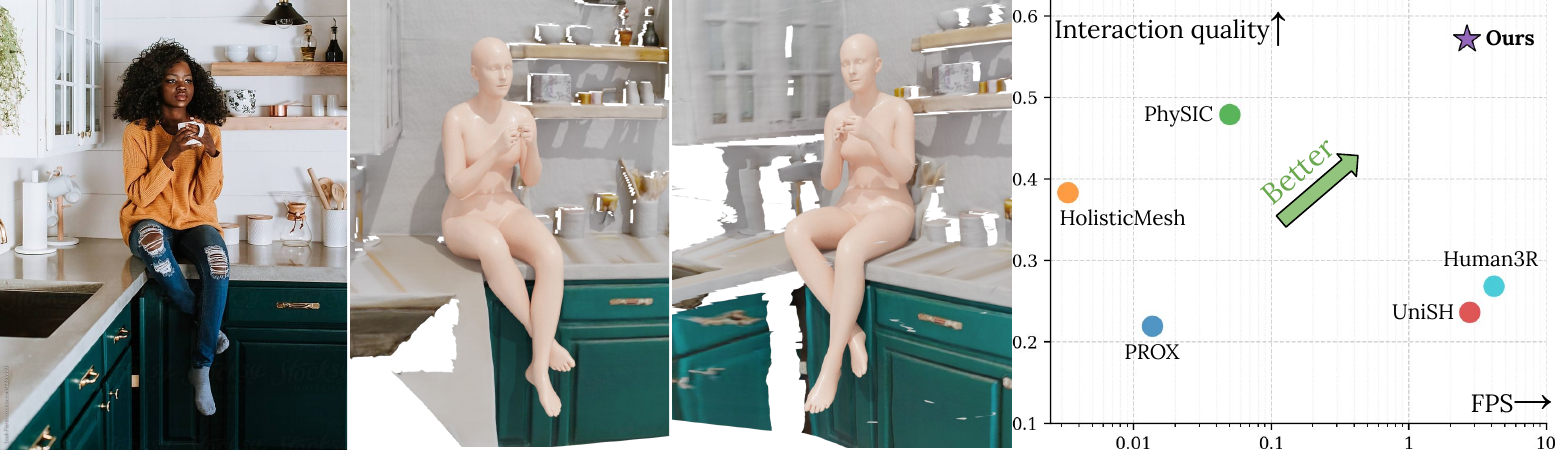}
    \vspace{-2em}
    \caption{\textbf{Fast, geometry-grounded human--scene reconstruction.} \emph{Left:} From a single RGB image, GRAFT jointly reconstructs the 3D human and scene with physically coherent interactions. \emph{Right:} GRAFT breaks the traditional speed--accuracy tradeoff, achieving high interaction accuracy at near feed-forward inference speeds, while existing methods typically gain one at the expense of the other.}
    \label{fig:teaser}

    \vspace{-2em}
\end{figure}

What is missing is a \emph{learned HSI prior}: a model that understands how humans physically relate to 3D scenes and can resolve floating and interpenetration in a single feed-forward pass—without costly test-time optimization. We therefore introduce \textbf{\modelName{}} (\textbf{G}eometric \textbf{R}efinement \textbf{A}nd \textbf{F}itting \textbf{T}ransformer). Our key insight is to \emph{amortize} geometry-based optimization into a feed-forward transformer: instead of minimizing handcrafted energy functions at test time, \modelName{} directly regresses the corrective trajectory—the \emph{Interaction Gradient}—from an implausible human-scene state to a physically grounded configuration.
Operating on explicit 3D inputs, \modelName{} supports two complementary modes: conditioned on image features it serves as a standalone HSI reconstruction system, and on geometry alone it acts as a universal plug-in prior that improves any existing method without retraining.

Concretely, \modelName{} encodes the interaction state into a set of \emph{HSI Tokens}—one per body joint, one per hand, and one for the full body—each grounded in the scene via \emph{Geometric Probes}: nearest-neighbor queries that capture metric distances, directions, and surface normals from the surrounding scene point cloud. A lightweight transformer refines these tokens via self-attention to model inter-limb dependencies (\eg, support-contact coupling between feet and torso) and \emph{Geometry-Aware Cross-Attention} to fuse geometric cues with image features, then decodes corrective parameter updates. Crucially, this is \emph{recurrent}: probes are recomputed after each update, giving the network direct physical feedback that progressively resolves penetrations and establishes contact.

Experimentally, \modelName{} improves contact F1 by up to 122\% over feed-forward baselines and matches optimization-based PhySIC at ${\sim}100{\times}$ lower runtime (0.2\,s vs.\ 20\,s; Fig.~\ref{fig:teaser}). In geometry-only mode it boosts Human3R's contact F1 by up to 44\% without retraining. Our method generalizes to in-the-wild images, including multi-person scenes and complex interactions, and is preferred in 64.8\% of trials in a three-way user study. Our contributions are as follows:
\begin{itemize}
    \item \textbf{GRAFT}: a learned HSI prior that amortizes geometry-based optimization into a feed-forward transformer, predicting corrective \emph{Interaction Gradients} through a recurrent refinement loop, explicitly reasoning about 3D HSI.
    \item \textbf{HSI Tokenization}: \emph{Geometric Probes} and \emph{Geometry-Aware Cross-Attention} anchor body-part queries directly to the scene point cloud and fuse metric 3D cues with image features, encoding the full interaction state in only 24 compact tokens.
    \item \textbf{Dual-mode architecture}: conditioned on image features, \modelName{} serves as a standalone HSI reconstruction system; on geometry alone, it acts as a universal plug-in prior that improves any existing method without retraining, boosting Human3R's contact F1 by up to 44\%.
\end{itemize}

\section{Related Works}
\label{sec:related_works} 

\subsection{3D Human-Scene Reconstruction}
\begin{wraptable}[12]{r}{0.49\linewidth}
    \centering
    \footnotesize
    \renewcommand{\arraystretch}{1.}
    \renewcommand{\tabcolsep}{2.5pt}
    \vspace{-22pt}
    \resizebox{\linewidth}{!}{
    \begin{tabular}{@{}l cccc@{}}
        \toprule
        \multirow{2}{*}{\textbf{Method}} & \multirow{2}{*}{\begin{tabular}{@{}c@{}}\textbf{HSI} \\ \textbf{Reasoning}\end{tabular}} & \multirow{2}{*}{\begin{tabular}{@{}c@{}}\textbf{Feed-} \\ \textbf{forward}\end{tabular}} & \multirow{2}{*}{\begin{tabular}{@{}c@{}}\textbf{Multi-} \\ \textbf{Human}\end{tabular}} & \multirow{2}{*}{\textbf{Runtime}} \\
        & & & & \\
        \midrule
        PROX~\cite{hassan2019resolving} & \cmark & \xmark & \xmark & 73 sec. \\
        HolisticMesh~\cite{weng2021holistic} & \cmark & \xmark & \xmark & 5 min. \\
        PhySIC~\cite{ym2025physic} & \cmark & \xmark & \cmark & 20 sec. \\
        Human3R~\cite{chen2025human3r} & \xmark & \cmark & \cmark & 0.24 sec. \\
        UniSH~\cite{li2026unish} & \xmark & \cmark & \xmark & 0.36 sec. \\
        \midrule
        \textbf{Ours} & \cmark & \cmark & \cmark & 0.2 sec. \\
        \bottomrule
    \end{tabular}}
    \vspace{-8pt}
    \caption{
    GRAFT achieves strong multi-human HSI reasoning at feed-forward speed. \textbf{Multi-Human} denotes reconstructing all humans and the scene in one shared metric frame, following PhySIC~\cite{ym2025physic}.
    }
    \label{tab:methods-comparison}
    \vspace{-2.0em}
\end{wraptable}

Despite strong progress in monocular human pose, surface, and avatar reconstruction~\cite{patel2024camerahmr, sarandi2024nlf, xue2023nsf, xue2024human3diffusion, xue2025gen3diffusion, xue2025infinihuman, xue2026georelight}, physically plausible human--scene interaction (HSI) reconstruction remains difficult. Optimization-based methods such as PROX~\cite{hassan2019resolving} model contact and interpenetration explicitly, but depend on static scene scans and expensive per-instance fitting. Follow-up methods (e.g., HolisticMesh~\cite{weng2021holistic,yi2022humanaware} and related RGB approaches) reduce static scene assumptions, while recent systems such as PhySIC~\cite{ym2025physic, liu2025joint,mueller2024hsfm} combine optimization with stronger depth/geometry priors. These pipelines preserve explicit interaction reasoning but remain slow and sensitive to initialization.
More recently, feed-forward video methods such as Human3R~\cite{chen2025human3r} and UniSH~\cite{li2026unish} achieve fast inference, but lack explicit HSI reasoning. As a result, they struggle with complex interactions (e.g., leaning on a wall, or lying on a couch) where multiple body parts must respect fine-grained surface constraints simultaneously.
GRAFT is designed to achieve the best of both worlds, combining feed-forward efficiency with explicit 3D HSI reasoning, faithfully recovering the full range of complex body--scene contacts while maintaining competitive runtime (Table~\ref{tab:methods-comparison}).

\subsection{Human-Scene Interaction Prior}
Early human-prior models such as Pose-NDF~\cite{tiwari2022posendf}, NRDF~\cite{he24nrdf}, and VPoser~\cite{pavlakos2019expressive} learn pose distribution/manifolds in isolation, without explicit scene grounding. Explicit HSI-prior methods~\cite{zhang2020place,hassan2021populating} condition on scene geometry, but follow a two-stage \emph{generate-then-optimize} paradigm: they first predict plausible contact maps or placement distributions, then run costly test-time optimization to fit candidate poses to each scene. This decouples scene sensing from pose correction, making the pipeline slow and fragile—the optimizer must bridge the gap between a coarse contact prediction and a full-body solution, often falling into local minima. Moreover, these methods encode the scene with dense body-centric representations (${\sim}4$K--$10$K parameters~\cite{zhang2020place,prokudin2019efficientl,hassan2021populating}), which risk overfitting to training-set geometry and scale poorly to partial observations.

GRAFT instead operates as a \emph{corrective} prior: given an existing (possibly wrong) human--scene configuration, it directly predicts interaction gradients that move the mesh toward a plausible state, unifying scene sensing and pose correction in a single forward pass. Its compact 24-token HSI representation (${\sim}500$ geometric parameters) captures the dominant interaction signal with far fewer dimensions, improving generalization and enabling the model to function as a geometry-only plug-in prior on top of other methods (Sec.~\ref{sec:exp}).

\subsection{Iterative Refinement}
Iterative residual refinement has seen widespread adoption in diverse vision systems~\cite{teed2020raft, teed2022droidslam, xu2025resplata, xie2026cari4d}.
This paradigm has naturally been adopted in human-centric tasks. 
ReFit~\cite{wang2023refit} refines SMPL estimation in a recurrent loop by reprojecting 3D keypoints onto 2D image features. WiLoR~\cite{potamias2025wilor} iteratively projects an initial mesh onto multi-scale image features to regress pose and shape residuals. Learned Vertex Descent (LVD)~\cite{corona2022learned} predicts 3D per-vertex displacements from localized feature projections, effectively amortizing an optimization descent.

GRAFT is the first work exploring the iterative refinement for 3D \emph{human--scene relationship}: geometric probes are recomputed from the updated mesh at every step, so as the body moves, distances, normals, and contact patterns shift—giving the network a direct physical gradient toward plausible interaction. Because this feedback is purely geometric, GRAFT can also operate without any visual features and still serve as a plug-in HSI prior atop any method (Table~\ref{tab:quantitative_comparison})—a capability no prior iterative method possesses.

\section{Method}
\label{sec:methods}

Given a single RGB image $\set{I}_h \in \real{H \times W \times 3}$ of humans in a scene, our goal is to coherently reconstruct both the 3D scene geometry $\scene_s \in \real{H \times W \times 3}$ and the human body meshes $\mathcal{M}_i=\operatorname{SMPLX}(\hparams_{i})$, with $i$ indexing the person instances.
Each per-human parameter vector is $\hparams=\{\pose, \trans, \shape\}$, with pose $\pose \in \real{165}$, translation $\trans \in \real{3}$, and shape $\shape \in \real{10}$.
In contrast to one-shot methods, \modelName{} iteratively predicts \textit{interaction gradients}---learned corrective parameter updates---by repeatedly probing the local scene geometry and attending to image tokens, progressively driving each mesh toward a geometry-grounded solution.

\begin{figure}[t!]
    \centering
    \includegraphics[width=1\linewidth]{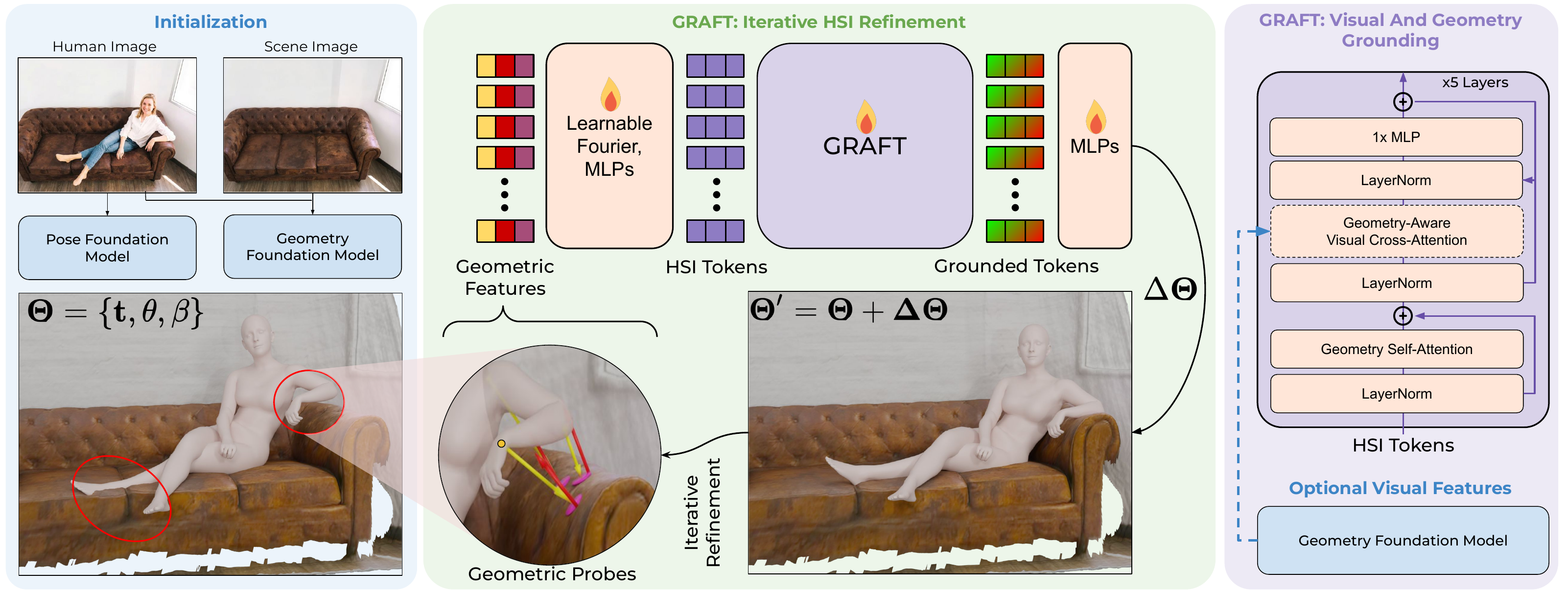}
    \vspace{-1em}
    \caption{\textbf{Overview of GRAFT.} \textit{Left:} Foundation models initialize human meshes (NLF~\cite{sarandi2024nlf}) and scene geometry (MapAnything~\cite{keetha2026mapanything}), often yielding misalignments. \textit{Center:} Geometric probes (nearest-neighbor scene points for body joints and body vertices) encode local contact cues (such as position and surface normals) into compact HSI tokens. GRAFT uses these tokens to predict iterative updates $\hparams' = \hparams + \dhparams$, correcting penetration and floating artifacts. \textit{Right:} GRAFT alternates geometric self- and visual cross-attention for HSI-prior-guided refinement, optionally fusing image features.}
    \label{fig:overview}
    \vspace{-2em}
\end{figure}

Our framework has two core components (Fig.~\ref{fig:overview}). We start from a coarse metric initialization of the human and scene using separate foundation models (Sec.~\ref{sec:HSI_initialization}), where the two predictions are typically still incoherent. We then apply \modelName{} (Sec.~\ref{sec:GRAFT_method}), an iterative refinement transformer that leverages compact HSI tokens and geometric probes capturing human–scene point cloud relations, with optional Geometry-Aware Cross-Attention grounded in image evidence. Training objectives are described in Sec.~\ref{sec:training_objectives}.

\subsection{HSI Initialization}
\label{sec:HSI_initialization}

To initialize iterative refinement, we use two foundation models: MapAnything~\cite{keetha2026mapanything} for scene geometry and features, and NLF~\cite{sarandi2024nlf} for initial human meshes $\mathcal{M}_i^\text{init}$. To recover occluded scene geometry behind humans, we first remove them from $\set{I}_h$ using OmniEraser~\cite{wei2025omni}, similar to~\cite{ym2025physic}, producing $\set{I}_s$. We then apply MapAnything to $\set{I}_s$ and $\set{I}_h$ jointly, yielding coherent scene pointmaps $\scene_s$ and $\scene_h$ in a shared camera frame; these features are reused in later stages. We recover metric scale using a monocular depth estimator~\cite{wang2025moge2}, following~\cite{lin2025depth3,li2024megasam}.

For coarse alignment, we project each mesh's head joint to 2D, retrieve the corresponding 3D scene point from $\scene_h$, and compute a per-human depth-ratio scale $s_i = p_{i,z}^{\text{scene}} / p_{i,z}^{\text{head}}$ to bring the mesh into the metric frame of $\scene_s$, yielding $\{\set{M}_i^{0}\}_{i=1}^{N_h}$. This provides only a coarse initialization; \modelName{} then independently refines each $\hparams_i$ using local geometric evidence and learned interaction priors, naturally supporting multi-person scenes with shared weights.

\subsection{Iterative Interaction Transformer}
\label{sec:GRAFT_method}
Our model takes as input the current interaction state defined by the scene pointmap $\scene_s$ and human parameters $\hparams=\{\pose, \trans, \shape\}$ and predicts the \emph{interaction gradient}: a corrective $\dhparams=\{\Delta\pose, \Delta\trans, \Delta\shape\}$ with an additional uniform scale $s$ to compensate residual metric-scale mismatch from initialization.
We predict $s$ explicitly since scale ambiguity dominates the mismatch; decoupling it from $\shape$ and $\trans$ simplifies the correction (Suppl.).
We propose parameter- and data-efficient architecture design by leveraging
(a) a \emph{compact HSI token representation} that captures localized 3D cues and (b) an \emph{efficient Geometry-Aware Cross-Attention mechanism} that grounds these cues in visual evidence.
In each recurrent step, the model closes the loop by re-sampling metric spatial evidence from the scene, enabling precise contact resolution that single-pass methods fail to achieve.

\subsubsection{HSI Tokenization with Geometric Probing}

Transformers are well suited for HSI refinement because they jointly model long-range body-part dependencies and can fuse geometry with image evidence through self-/cross-attention.
We encode the current interaction state $(\scene_s,\hparams)$ as a compact set of \textit{HSI tokens} and keep this representation low-dimensional for both \emph{generalization} and \emph{efficiency}: it discourages overfitting to high-dimensional noise, encourages learning the underlying interaction manifold, and reduces quadratic attention cost.
Our tokenization aggregates local geometric cues around the body into a fixed, small number of tokens (instead of raw points or dense maps), using ${\sim}500$ parameters versus ${\sim}4\mathrm{K}$ for BPS (basis-point-set distance) representations~\cite{zhang2020place,prokudin2019efficientl} and $\sim 10\mathrm{K}$ for POSA (dense body-surface contact) representations~\cite{hassan2021populating}.

Each token is built from one or more \emph{geometric probes} that capture the local 3D human–scene relationship at a specific body location. Each probe is a spatial query anchored at a 3D point $\mathbf{p}$ (e.g., a joint or surface vertex). Since signed distance fields are ill-defined for partial observations such as pointmaps, for each $\mathbf{p}$, we retrieve the closest scene point $\mathbf{p}^*$ and compute the relative offset $\mathbf{v}_s = \mathbf{p}^* - \mathbf{p}$ together with the corresponding scene normal $\mathbf{n}^*$. We also express $\mathbf{p}$ in body-relative coordinates. Fused with visual features via cross-attention, these signals enable reasoning about local contact, penetration, and support. All geometric vectors $\{\mathbf{v}_s, \mathbf{n}^*, \mathbf{p}\}$ are encoded using learnable Fourier features~\cite{li2021learnable}.

We define 24 fixed HSI tokens for each human: 21 body joint tokens (non-root joints), 2 hand tokens (left/right), and 1 full-body token. This layout mirrors SMPL-X's output groups, so each token gathers the geometric and visual context relevant to the parameters it predicts.
Each token concatenates (i) geometric probe features and (ii) token-specific SMPL-X parameter context, then projects them to a shared embedding space via an MLP. In compact form,
\begin{equation*}
\mathbf{z}_k^{t}=\phi_k\!\left([\mathbf{g}_k^{t};\mathbf{p}_k^{t}]\right),\qquad
\mathbf{Z}^{t}=\{\mathbf{z}_k^{t}\}_{k=1}^{24},
\end{equation*}
where $\mathbf{g}_k^{t}$ are geometric probe encodings and $\mathbf{p}_k^{t}$ are token-specific SMPL-X parameter features.
Each \emph{body joint token} carries one probe at its joint together with the joint's 6D rotation;
each \emph{hand token} aggregates five distal-joint probes and the corresponding finger rotations;
and the \emph{full-body token} aggregates 27 surface probes at fixed vertices uniformly sampled from the SMPL-X template mesh, together with global orientation, translation, and shape parameters.
Tokenization details and the MLP architecture are provided in the supplementary.

\subsubsection{Visual Anchors and Geometry-Aware Attention}
To ground each token in image evidence, we use two MapAnything feature streams: a scene stream from the human-removed image $\set{I}_s$ and an interaction stream from the original image $\set{I}_h$. For each token, we define one or more \emph{visual anchors}: 3D body points $\mathbf{a} \in \real{3}$ projected to both images by $\mathbf{u}=\pi(\mathbf{a})$. Around each $\mathbf{u}$, we sample multi-scale MapAnything features from both streams and concatenate them as the token context. Body tokens use one anchor per body joint, hand tokens use one anchor per hand (mean of distal joints), and the full-body token uses 27 surface anchors. We sample $3\times3$ MapAnything token neighborhoods around each body/hand anchor and $1\times1$ (single-token) features for the 27 full-body anchors to preserve efficiency.

We use a 5-layer transformer with standard multi-head attention, alternating: (i) self-attention among HSI tokens and (ii) constrained cross-attention to sampled visual features. In self-attention, geometry-aware tokens exchange global interaction context (e.g., support-contact dependencies between limbs and torso). In cross-attention, token $k$ attends only to features sampled at its own anchors, yielding a sparse connectivity pattern that tightly couples geometric probes with local visual evidence.
Let $\mathbf{V}^{t}=\{\mathbf{v}_k^{t}\}_{k=1}^{24}$ denote anchor features sampled from both streams. One attention block can be written as follows, with $\widehat{\mathbf{Z}}^{t}$ denoting the geometry-grounded interaction state:
\begin{equation*}
\widetilde{\mathbf{Z}}^{t}=\texttt{Self-Attn}(\mathbf{Z}^{t}),\qquad
\widehat{\mathbf{Z}}^{t}=\texttt{Cross-Attn}(\widetilde{\mathbf{Z}}^{t},\mathbf{V}^{t}),
\end{equation*}

\subsubsection{Iterative Interaction Refinement}

At iteration $t$, we decode step-wise updates directly from the geometry-grounded state $\widehat{\mathbf{Z}}^{t}$. We decode 6D rotations for the 21 body joints from the body tokens, 6D rotations for the 15 joints in each hand from the hand tokens, and global orientation from the full-body token, together with shape and translation updates. In compact form,
\begin{equation*}
(\dhparams^{t}, s^{t})=\Psi(\widehat{\mathbf{Z}}^{t}),\qquad
\hparams^{t+1}=\hparams^{t}+\dhparams^{t}.
\end{equation*}
Lightweight MLP heads implement $\Psi$, predicting $\Delta\pose^{t}$ (body, hand joint, and global rotations), $\Delta\trans^{t}$, and $\Delta\shape^{t}$, while a separate head on the full-body token predicts a uniform scale $s^{t}$ applied to the vertices.
We maintain the rotation representation in 6D for stable updates, then convert to axis-angle for the SMPL-X forward pass to produce the refined mesh $\mathcal{M}^{t}$.

This updated state is fed back into the next iteration: we recompute geometric probes, resample visual anchors, and re-run the transformer with shared weights across timesteps. Over $T$ iterations, the meshes progressively reduce interpenetration and improve contact and alignment with the scene geometry. We describe the supervision used to train this iterative trajectory in Sec.~\ref{sec:training_objectives}.

\subsubsection{Fast Differentiable Scale Update}
In addition to pose, translation, and shape updates, GRAFT predicts a per-step uniform scale $s^{t}$ applied to mesh vertices. During training, naively re-fitting SMPL-X shape coefficients at every step to match scaled vertices is prohibitively expensive. We therefore use a closed-form linear approximation: because shape blendshapes act linearly on vertices, a uniform scaling can be absorbed into shape parameters as
\begin{equation*}
\shape_s = s \cdot (\shape + \mathbf{c}) - \mathbf{c},
\end{equation*}
where $\mathbf{c}$ is the least-squares projection of the mean body template into $\shape$-space, computed once offline and fixed for a given body model (SMPL-X). Translation scales directly and rotations are unchanged, replacing complex per-step fitting with simple vector arithmetic. This yields an efficient, fully differentiable update for multi-step rollout supervision. Full derivation is provided in the Suppl..

\subsection{Training Objectives}
\label{sec:training_objectives}

We train \modelName{} on paired human--scene data with ground-truth SMPL-X annotations, supervising the full iterative trajectory by applying losses at every refinement step $t$ rather than only at the final output. This per-step supervision exposes the network to a richer set of off-manifold states and encourages monotonic improvement across iterations.

\boldparagraph{Training queries.} At each iteration we sample a mixture of NLF-initialized queries (realistic misalignments) and ground-truth parameters with random perturbations. This teaches the model to both (i) correct large initialization errors and (ii) remain stable near the optimum by predicting near-zero updates when the input is already well aligned.

\boldparagraph{Per-step losses.} At each step $t$, we combine three complementary terms:
\begin{equation*}
\mathcal{L} = \sum_{t=1}^{T} \big( \lambda_{p}\,\lVert \pose^{t} - \pose^{*} \rVert_{2}^{2} + \lambda_{v}\,\lVert \mathcal{V}^{t} - \mathcal{V}^{*} \rVert_{2}^{2} + \lambda_{n}\,\lVert \widetilde{\mathcal{V}}^{t} - \widetilde{\mathcal{V}}^{*} \rVert_{2}^{2} \big),
\end{equation*}
where $\pose^{t}$ denotes the predicted 6D joint rotations, $\mathcal{V}^{t}$ are camera-relative vertices (capturing the joint effect of $\trans$ and scale $s$), and $\widetilde{\mathcal{V}}^{t}$ are mean-normalized vertices that isolate shape supervision from global placement. The rotation loss directly penalizes articulation errors in the continuous 6D space, while the two vertex losses decouple global positioning from body proportions.
Notably, we do \emph{not} employ explicit contact or interpenetration losses. Our geometric probes already encode local human--scene distances and normals at every step, providing dense spatial evidence from which contact and non-penetration constraints emerge implicitly through regression to ground-truth poses. Moreover, such geometric losses are ill-defined for the partial, single-view point clouds we operate on, where occluded surfaces introduce spurious nearest-neighbor correspondences. We find that rich per-step geometric input combined with trajectory supervision is sufficient to learn plausible interactions without auxiliary HSI losses (Sec.~\ref{sec:exp}).

\boldparagraph{Training strategy.} We stabilize iterative training with two techniques: \emph{curriculum rollout}, which gradually increases the number of supervised refinement steps from zero to $T$, and \emph{visual-anchor dropout}, which randomly drops per-token visual neighborhoods from both scene and interaction streams to prevent over-reliance on appearance. All hyperparameters (loss weights $\lambda_p,\lambda_v,\lambda_n$, mixture ratios, and training schedules) are provided in Sec.~\ref{sec:exp}.

\begin{figure}[t]
    \centering
    \includegraphics[width=\linewidth]{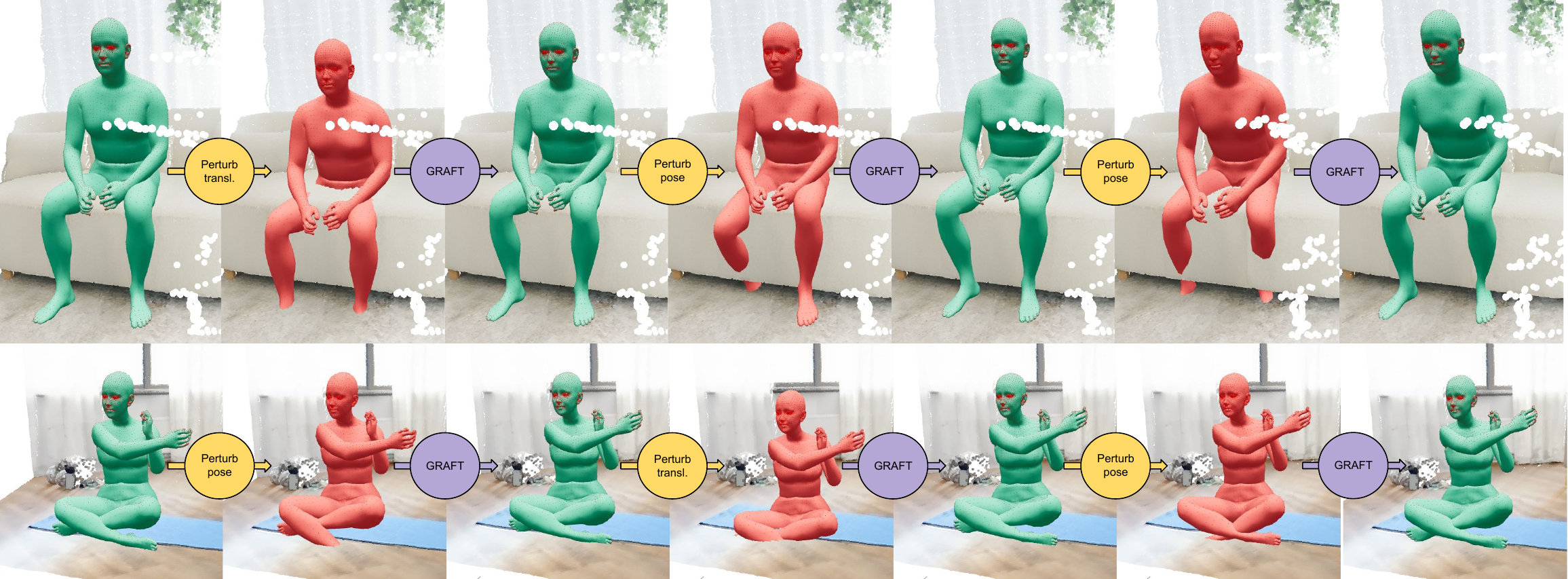}
    
    \caption{\textbf{GRAFT as a learned HSI prior.} Starting from an initial state (green mesh), we apply translation and pose perturbations (red); after each perturbation, GRAFT---operating with \emph{no visual features}---projects the state back to a geometrically valid human--scene interaction (green). Our geometric probes encode contact and penetration cues that drive each correction step. We refer readers to the supplementary video for a comprehensive visualization.}
    \label{fig:refinement_prior}
    \vspace{-2em}
\end{figure}

\section{Experiments}
\label{sec:exp}
\subsection{Implementation Details}
\boldparagraph{Datasets.} We train GRAFT on a pseudo-labeled HSI dataset InHabitants~\cite{inhabit2026}, comprising $75$k images and $97$k human instances.
Other datasets do not offer both realistic RGB and accurate interaction annotations: RICH~\cite{huang2022capturing} offers accurate motion captures and 3D scene scans, but is limited to $<$15 indoor/outdoor scenes, PROX-D~\cite{hassan2019resolving} is captured on 12 indoor scenes and uses optimization pseudo-labels with unnatural and inaccurate poses, BEDLAM~\cite{black2023bedlam} lacks realistic interactions (humans float above furniture), and HUMANISE~\cite{wang2022humanise} provides no photorealistic RGB.
GRAFT is inherently bounded by pseudo-label quality; scaling to richer data is a promising future direction.

\boldparagraph{Optimization.} We train with Adam for $215$k iterations on a single H100 GPU. The peak learning rate is $1\times10^{-4}$, with linear warmup followed by cosine decay.

\boldparagraph{Training queries and perturbations.} Following Sec.~\ref{sec:training_objectives}, each training step uses a mixture of NLF-initialized queries and GT-based queries. For GT-based queries, we keep clean GT states with probability $0.2$, and with probability $0.8$ we add Gaussian perturbations to improve robustness around the interaction manifold: translation noise with standard deviation $0.1$ m, SMPL-X shape ($\beta$) noise with standard deviation $0.03$, pose rotation noise with standard deviation $7^\circ$, and global orientation noise with standard deviation $3^\circ$.

\boldparagraph{Rollout training and scaling.} We supervise all refinement steps and train with curriculum rollout, increasing from single-step supervision to full $T{=}3$-step supervision. Full rollout supervision starts after the first $10$k training iterations. As in Sec.~\ref{sec:GRAFT_method}, refinement weights are shared across iterations. we set $T{=}3$ for both training and inference, as additional iterations yield only marginal gains (Fig.~\ref{fig:f1_runtime_halfcol}). During training, we also apply random global scale augmentation by sampling a factor from $[0.85, 1.15]$ and use our fast differentiable scale update (Sec.~\ref{sec:GRAFT_method}) to keep this operation efficient. We apply visual-anchor dropout with probability $0.35$ per token.

\begin{figure}[t]
    \centering
    \includegraphics[width=1\linewidth]{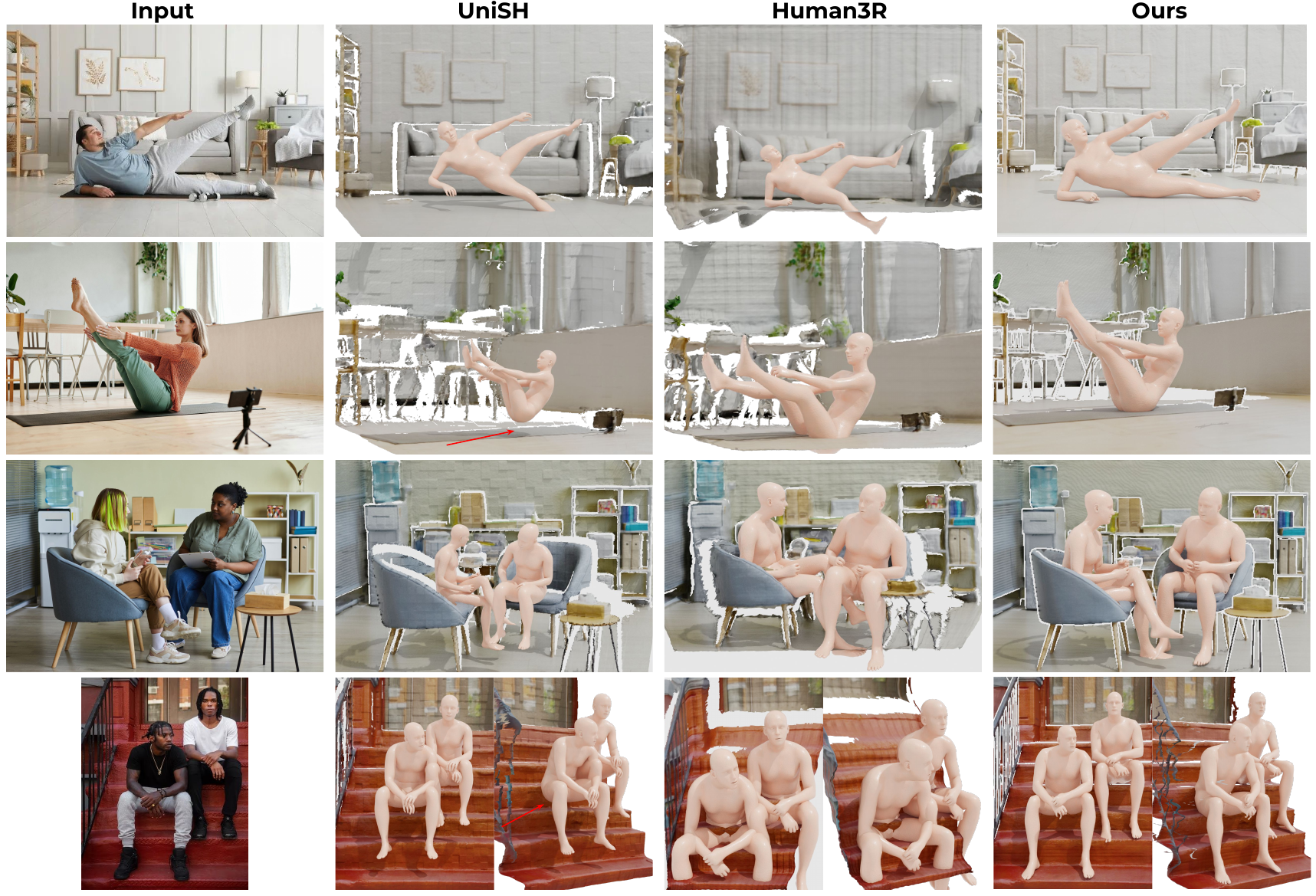}
    \vspace{-2em}
    \caption{
    \textbf{In-the-wild qualitative comparison.} We compare GRAFT against feed-forward methods, UniSH~\cite{li2026unish} and Human3R~\cite{chen2025human3r} on unconstrained internet images. While prior approaches localize humans in the scene, they lack explicit human--scene interaction modeling, often resulting in weak support, hovering or penetration. In contrast, GRAFT recovers physically coherent interactions with scene-consistent contact. We refer readers to the supplementary for more results.}
    \label{fig:qual_results}
    \vspace{-2em}
\end{figure}

\subsection{Evaluation Protocol}
\boldparagraph{Baselines.} We train our method from scratch and compare against both optimization based and feed-forward baselines. Specifically, we evaluate PROX and PhySIC (test-time optimization) as well as Human3R and UniSH (feed-forward methods trained on synthetic/large-scale video data). Human3R and UniSH are video-based models and, in the absence of dedicated single-image HSI reconstruction baselines, they are the closest feed-forward baselines for comparison. For fair comparison, and because interaction evaluation requires complete scene geometry, we pass the inpainted scene image $\set{I}_s$ through the depth backbones used by the baselines (CUT3R~\cite{wang2025continuous} and Pi3~\cite{wang2025p3}) and robustly align scene points to their incomplete predictions, so all methods are evaluated under the same complete-geometry protocol.

\boldparagraph{Metrics.} In addition to standard contact precision/recall/F1, we introduce two geometric interaction metrics that provide a continuous measure of human--scene relative positioning. Contact precision/recall/F1 remains our hard-constraint evaluation, while the new metrics quantify geometric consistency continuously. Using the PROX contact-vertex set~\cite{hassan2019resolving}, for each contact vertex we compute the vector from the human vertex to its nearest scene point in both prediction and ground truth. We then report (i) \textbf{V2S} (Vector-to-Scene): the weighted mean Euclidean error between predicted and ground-truth vectors (reported in mm), and (ii) \textbf{D2S} (Direction-to-Scene): the weighted mean angular error between these vectors (reported in degrees). Per-vertex weights are density-aware, so high-vertex-density regions (e.g., hands) do not dominate the error term.

Following the evaluation protocol of PhySIC, we report quantitative results on RICH-100 and PROX-test, both of which provide ground-truth human and scene scans. In addition, we present qualitative results on PiGraphs and demonstrate strong generalization on curated in-the-wild internet images.

\subsection{Qualitative Evaluation}

\begin{figure}[t]
    \centering
    \includegraphics[width=1\linewidth]{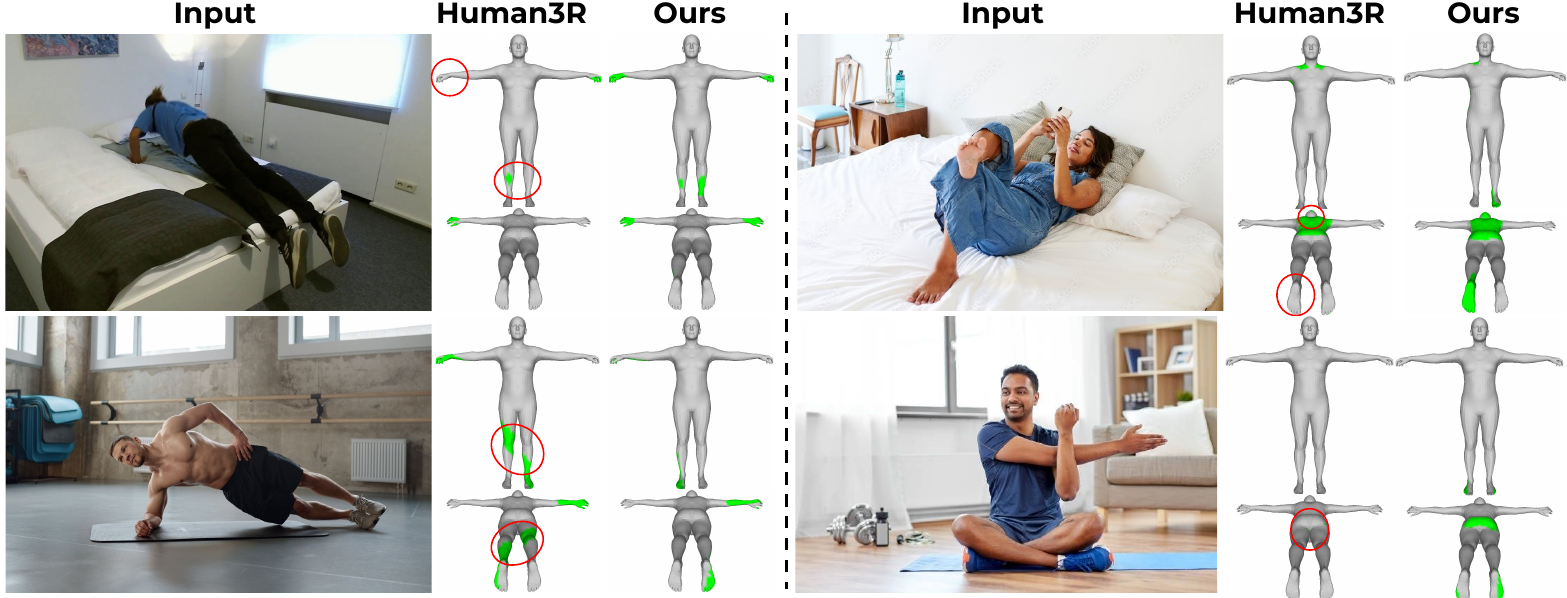}
    \caption{\textbf{Contact maps from geometry.} By reconstructing accurate 3D human--scene interaction, GRAFT derives contact directly by spatial proximity, yielding sharp and reliable contact regions.}
    \label{fig:qualitative_contact}
    \vspace{-1em}
\end{figure}

Compared with UniSH and Human3R, GRAFT produces more realistic human--scene interactions, especially at support and contact regions (e.g., feet--floor and body--furniture interfaces; Fig.~\ref{fig:qual_results}). While these baselines often localize the human reasonably in the scene, they frequently fail to capture fine interaction structure, leading to weak support, hover artifacts, or penetration near local geometry. In contrast, our iterative refinement predicts interaction gradients from geometry-grounded local probes and updates parameters incrementally rather than regressing the full pose from scratch, enabling reliable reconstruction even for challenging articulated poses. Because refinement is applied per human on shared scene geometry with shared weights, GRAFT naturally handles multi-person scenes while preserving coherent relative placement, and generalizes strongly to in-the-wild images despite training on a comparatively small pseudo-labeled dataset. As shown in Fig.~\ref{fig:qualitative_contact}, accurate contact maps emerge directly from the 3D reconstruction via spatial proximity, without contact-specific post-processing.

Fig.~\ref{fig:refinement_prior} provides a qualitative illustration of the learned HSI prior. Beginning from an initial reconstruction, we apply successive perturbations---first to global translation, then to body pose---and run GRAFT with no visual features. In both cases, GRAFT projects the perturbed state back onto the manifold of geometrically valid human--scene interactions, recovering plausible contact and support from geometry alone. This confirms that geometric probes are sufficient to capture a strong, generalizable interaction prior, and explains why the geometry-only mode transfers effectively as a plug-in prior to other feed-forward methods. We refer readers to the supplementary video for a more comprehensive visualization of this behavior across diverse scenes and perturbation magnitudes.

\subsection{Quantitative Evaluation}

\begin{table}[t]
\centering
\footnotesize
\renewcommand{\arraystretch}{1.}
\renewcommand{\tabcolsep}{2.5pt}
\resizebox{0.995\linewidth}{!}{
    \begin{tabular}{@{}c@{\hspace{6pt}}l c ccc  ccc@{}}
    \toprule
    \multirow{2}{*}{\begin{tabular}{@{}c@{}}\textbf{Test} \\ \textbf{Data}\end{tabular}} & \multirow{2}{*}{\textbf{Method}} & \multirow{2}{*}{\begin{tabular}{@{}c@{}}\textbf{Feed-} \\ \textbf{forward}\end{tabular}} & \multicolumn{3}{c}{\textbf{Human Pose Metrics} $\downarrow$} & \multicolumn{3}{c}{\textbf{Contact Metrics} $\uparrow$} \\
    \cmidrule(lr){4-6} \cmidrule(lr){7-9} 
    & & & {PA-MPJPE} & {V2S} & {D2S} & {Precision} & {Recall} & {F1 score} \\
    \midrule
    \multirow{5}{*}{\rotatebox[origin=c]{90}{\textit{RICH-100}}} & PROX~\cite{hassan2019resolving}         & \xmark & 120.24   & 538.67  & 79.38  & 0.069   & 0.253  & 0.108 \\ 
    & PhySIC~\cite{ym2025physic}       & \xmark & {{46.50}} & {237.69} & {41.65} & {{0.538}} & {{0.695}} & {{0.606}} \\ 
    \cmidrule(lr){2-9}
    & UniSH~\cite{li2026unish}        & \cmark & {{69.30}} & {{326.80}} & {{36.23}} & {{0.329}} & {{0.356}} & {{0.342}} \\ 
    & Human3R~\cite{chen2025human3r}      & \cmark & {{48.84}} & {274.05} & {{44.68}} & {{0.282}} & {{0.531}} & {{0.368}} \\ 
    & Human3R + Ours      & \cmark & {48.84} & {240.83} & {40.68} & {0.378} & {{0.655}} & {{0.479}} \\
    & \textbf{Ours}         & \cmark & {{\textbf{46.32} {\scriptsize\textcolor{Fgreen}{\textbf{($\downarrow$5.2\%)}}}}} & {{\textbf{224.66} {\scriptsize\textcolor{Fgreen}{\textbf{($\downarrow$18.0\%)}}}}} & {{\textbf{34.60} {\scriptsize\textcolor{Fgreen}{\textbf{($\downarrow$22.6\%)}}}}} & {{\textbf{0.473} {\scriptsize\textcolor{Fgreen}{\textbf{($\uparrow$67.7\%)}}}}} & {{\textbf{0.743} {\scriptsize\textcolor{Fgreen}{\textbf{($\uparrow$39.9\%)}}}}} & {{\textbf{0.578} {\scriptsize\textcolor{Fgreen}{\textbf{($\uparrow$57.1\%)}}}}} \\
    \midrule[\heavyrulewidth]
    \multirow{6}{*}{\rotatebox[origin=c]{90}{\textit{PROX-test}}} & PROX~\cite{hassan2019resolving}         & \xmark & 76.98   &  224.49  &  59.60 & 0.458   & 0.144  & 0.219 \\
    & HolisticMesh~\cite{weng2021holistic} & \xmark & 81.30   & {170.64} & {54.48} & 0.427   & 0.348  & 0.383 \\
    & PhySIC~\cite{ym2025physic}       & \xmark & {{44.17}} & 148.05  & 47.11  & {{0.550}} & {{0.424}} & {{0.479}} \\
    \cmidrule(lr){2-9}
    & UniSH~\cite{li2026unish}        & \cmark & {59.02} & {256.38} & {60.34} & {0.362} & {0.175} & {0.236} \\
    & Human3R~\cite{chen2025human3r}      & \cmark & {59.25} & {200.35} & {59.36} & {0.228} & {0.324} & {0.268} \\
    & Human3R + Ours      & \cmark & {59.25} & {187.04} & {54.24} & {0.350} & {0.434} & {0.387} \\
    & \textbf{Ours}         & \cmark & {{\textbf{49.73} {\scriptsize\textcolor{Fgreen}{\textbf{($\downarrow$16.1\%)}}}}} & {{\textbf{184.36} {\scriptsize\textcolor{Fgreen}{\textbf{($\downarrow$8.0\%)}}}}} & {{\textbf{50.92} {\scriptsize\textcolor{Fgreen}{\textbf{($\downarrow$14.2\%)}}}}} & {{\textbf{0.556} {\scriptsize\textcolor{Fgreen}{\textbf{($\uparrow$143.9\%)}}}}} & {{\textbf{0.638} {\scriptsize\textcolor{Fgreen}{\textbf{($\uparrow$96.9\%)}}}}} & {{\textbf{0.594} {\scriptsize\textcolor{Fgreen}{\textbf{($\uparrow$121.6\%)}}}}} \\
    \bottomrule
    \end{tabular}
}
\caption{\textbf{Quantitative comparison on RICH-100 and PROX-test.} GRAFT consistently improves interaction quality over recent feed-forward baselines, with the strongest gains on human-scene metrics and the best PA-MPJPE among feed-forward methods. Gains are computed w.r.t.\ Human3R. Compared with optimization-heavy methods, GRAFT achieves similar interaction quality with much faster inference (\mbox{Tab.~\ref{tab:methods-comparison}}). \emph{Human3R\,+\,Ours} uses GRAFT's geometry-only (w/o-visual) variant.}
\label{tab:quantitative_comparison}
\vspace{-2em}
\end{table}

As shown in Table~\ref{tab:quantitative_comparison}, GRAFT consistently improves interaction quality over recent feed-forward baselines on both PROX-test and RICH-100, with the largest gains in contact and geometric interaction metrics, while keeping PA-MPJPE competitive. The reusable-refinement effect is also clear: applying GRAFT in geometry-only mode (no visual features) on top of Human3R improves interaction quality on both datasets — on PROX-test, F1 improves from $0.268\rightarrow0.387$ (with V2S $200.35\rightarrow187.04$), and on RICH-100, F1 improves from $0.368\rightarrow0.479$ (with V2S $274.05\rightarrow240.83$) — demonstrating that GRAFT functions as a plug-in HSI refinement prior (see Fig.~\ref{fig:refinement_prior}). Compared with optimization-heavy methods, GRAFT remains slightly behind the strongest optimization baseline on some metrics, but preserves feed-forward runtime (Table~\ref{tab:methods-comparison}). The two benchmarks also differ in domain (PROX: indoor/residential, frequent occlusions; RICH: mixed indoor/outdoor, large mostly-flat scenes), and as \modelName{}'s training data (InHabitants) is closer to PROX, this partly explains the dataset-dependent trends. GRAFT is also robust to initialization noise: under Gaussian depth-shift perturbations it recovers $50$--$60\%$ of the induced MPJPE error and keeps contact F1 above feed-forward baselines' noise-free F1 even at extreme noise (Suppl.).

\begin{wrapfigure}{r}{0.5\textwidth}
    \vspace{-\intextsep}
    \centering
    \includegraphics[width=\linewidth]{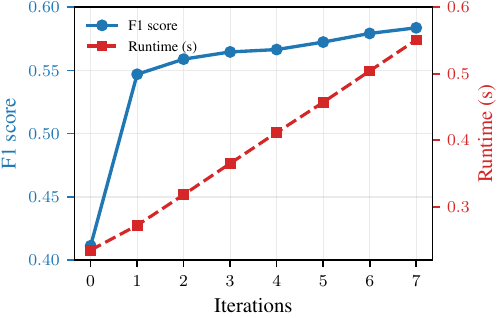}
    \caption{\textbf{Refinement iterations: quality--runtime trade-off.} Contact F1 improves sharply at the first refinement iteration, while additional iterations provide marginal gains. Runtime increases approximately linearly with iteration count.}
    \label{fig:f1_runtime_halfcol}
    \vspace{-2em}
\end{wrapfigure}

Fig.~\ref{fig:f1_runtime_halfcol} shows the convergence behavior across refinement iterations. We observe the most significant F1 jump at the first iteration, indicating that a single update already corrects most coarse interaction errors. Further iterations continue to improve performance, but with diminishing returns, yielding marginal gains per step while runtime grows steadily.

We conduct a perceptual study comparing GRAFT, UniSH, and Human3R via an online survey. Each participant evaluates a fresh random sample of 20 images (2 PROX-D, 3 PiGraphs, 15 of ${\sim}100$ in-the-wild) as interactive 3D reconstructions, selecting the most plausible human--scene interaction; method order is randomized and model names are hidden. Across $n{=}53$ participants (912 responses), GRAFT was chosen 591 times ($64.8\%$), well above the $33\%$ chance baseline (Fig.~\ref{fig:user_study}).

\subsection{Ablation Study}
\label{sec:ablation}
\begin{table}[t]
\newsavebox{\ablationbox}
\savebox{\ablationbox}{%
\begin{minipage}{0.74\linewidth}

\centering
\footnotesize
\renewcommand{\arraystretch}{1.}
\renewcommand{\tabcolsep}{2.5pt}
\resizebox{\linewidth}{!}{
    \begin{tabular}{@{}l ccc ccc@{}}
    \toprule
    \multirow{2}{*}{\textbf{Method}} & \multicolumn{3}{c}{\textbf{Human Pose Metrics} $\downarrow$} & \multicolumn{3}{c}{\textbf{Contact Metrics} $\uparrow$} \\
    \cmidrule(lr){2-4} \cmidrule(lr){5-7} 
    & {PA-MPJPE} & {V2S} & {D2S} & {Precision} & {Recall} & {F1 score} \\
    \midrule
    \multicolumn{7}{l}{\textit{RICH-100 Dataset}} \\ 
    Human3R$^*$ & 54.78 & 334.73 & 47.86 & 0.310 & 0.444 & 0.365 \\ 
    Ours (Init.) & 43.51 & 240.30 & 37.15 & 0.349 & 0.499 & 0.411 \\ 
    w/o rollout & \cellcolor{heatOrange}51.30 & \cellcolor{heatRed}274.83 & \cellcolor{heatRed}50.74 & \cellcolor{heatRed}0.270 & \cellcolor{heatOrange}0.532 & \cellcolor{heatRed}0.358 \\ 
    w/o geometric feat. & 48.76 & \cellcolor{heatOrange}273.33 & \cellcolor{heatOrange}42.19 & \cellcolor{heatOrange}0.315 & \cellcolor{heatRed}0.509 & \cellcolor{heatOrange}0.389 \\ 
    w/o scale pred. & \cellcolor{heatYellow}50.29 & \cellcolor{heatYellow}239.45 & 36.45 & \cellcolor{heatYellow}0.420 & \cellcolor{heatYellow}0.642 & \cellcolor{heatYellow}0.508 \\ 
    w/o GT-training & \cellcolor{heatRed}65.56 & 231.76 & \cellcolor{heatYellow}37.56 & 0.425 & 0.752 & 0.544 \\ 
    w/o visual feat. & 49.60 & 223.00 & 33.99 & 0.449 & 0.777 & 0.569 \\ 
    \textbf{Ours}  & 49.30 & 222.54 & 36.80 & 0.441 & 0.784 & 0.565 \\
    \cmidrule(lr){1-7}
    + V2S loss & 50.69 & 222.32 & 33.61 & 0.480 & 0.790 & 0.597 \\
    + contact \& penetr. loss & 50.04 & 221.65 & 34.26 & 0.460 & 0.728 & 0.563 \\
    
    \midrule[\heavyrulewidth]
    \multicolumn{7}{l}{\textit{PROX-test Dataset}} \\
    w/o visual feat. & 54.99 & 189.12 & 52.02 & 0.534 & 0.512 & 0.523 \\
    \textbf{Ours} & 54.83 & 188.71 & 51.67 & 0.526 & 0.627 & 0.572 \\
    \cmidrule(lr){1-7}
    + V2S loss & 53.75 & 186.21 & 51.19 & 0.544 & 0.584 & 0.563 \\
    + contact \& penetr. loss & 53.57 & 186.04 & 50.42 & 0.558 & 0.629 & 0.591 \\
    
    \bottomrule
    \end{tabular}
}

\end{minipage}}
\usebox{\ablationbox}%
\hfill
\begin{minipage}[c][\dimexpr\ht\ablationbox+\dp\ablationbox\relax][c]{0.24\linewidth}
\vspace*{1.3em}
\centering
\includegraphics[width=\linewidth]{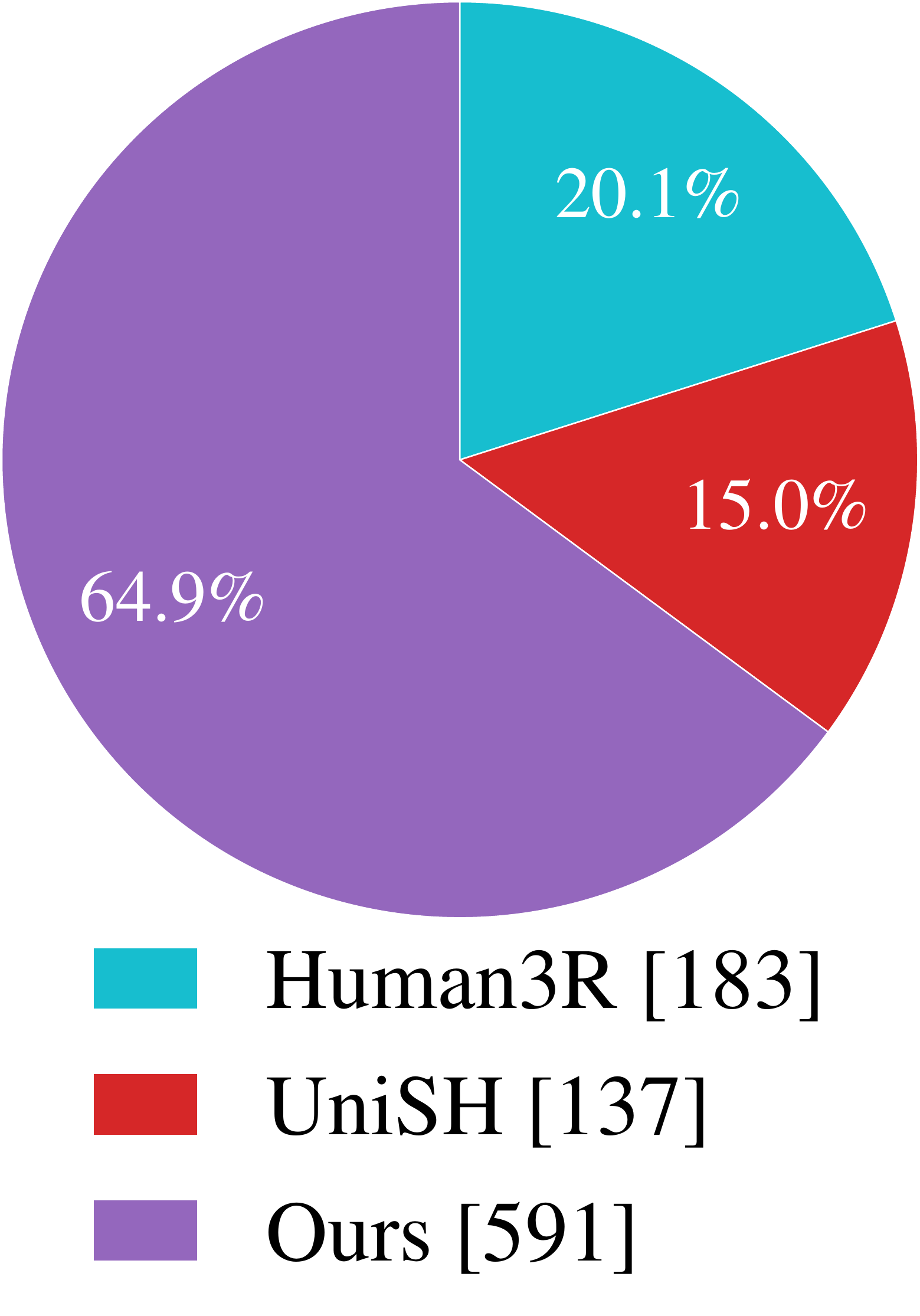}

\end{minipage}
\vspace{5pt}
\caption{\textit{(Left)}~Ablation study on RICH-100, with an additional visual-feature ablation on PROX. The bottom rows of each block (\emph{+\,V2S}, \emph{+\,contact\,\&\,penetr.}) add explicit HSI losses on top of \modelName{}. All ablations are trained for $120$k steps. Heatmap shading indicates degradation severity among design-choice variants (\colorbox{heatRed}{worst}, \colorbox{heatOrange}{second}, \colorbox{heatYellow}{third}). $^*$Human3R retrained on the same training data as GRAFT. \textit{(Right)}~User study: GRAFT is preferred by participants in $64.8\%$ of trials (912 responses, 53 participants), nearly twice the $33\%$ chance rate.}
\label{tab:ablations}
\label{fig:user_study}
\vspace{-2em}
\end{table}

Table~\ref{tab:ablations} analyzes each design choice by measuring the degradation it causes relative to the full model; heatmap shading highlights the severity of each drop. As a data-control, Human3R retrained on our pseudo-labeled set still lags clearly behind GRAFT, confirming our gains are architectural. Notably, \emph{Ours (Init.)} achieves a lower PA-MPJPE than the full model. This reflects an expected trade-off: PA-MPJPE is Procrustes-aligned and scene-agnostic, whereas GRAFT is supervised with pseudo-GT that intentionally adjusts articulation for scene consistency, diverging from mocap GT in pose space. The moderate PA-MPJPE increase ($43.51{\rightarrow}49.30$, on par with Human3R's $48.84$) is accompanied by large interaction gains (F1 $0.411{\rightarrow}0.565$, V2S $240{\rightarrow}223$\,mm), confirming that GRAFT learns scene-grounded rather than scene-agnostic articulation; this gap would shrink with higher-fidelity training data. \emph{Rollout supervision} and \emph{geometry groun\-ded features} are the two most critical components: removing either causes the largest drops across interaction metrics (rollout: F1 $0.565\rightarrow0.358$; geometric features: F1 $0.565\rightarrow0.389$, V2S $222.54\rightarrow273.33$). Without rollout, the model cannot learn multi-step correction trajectories; without scene-grounded tokens, it loses direct evidence of contact and penetration. \emph{Explicit scale prediction} also helps (removing it drops F1 $0.565\rightarrow0.508$): it decouples scale from translation, so the model need not absorb metric-scale errors through coupled translation and $\beta$ updates, where the $\beta$--scale mapping is nonlinear. \emph{GT-training} queries are essential for stability near the optimum: without them the model over-corrects already-good initializations (PA-MPJPE $49.30\rightarrow65.56$). Finally, \emph{visual features} have a limited effect on RICH but provide a clearer gain on PROX, mainly through higher recall. High-dimensional visual tokens risk overfitting on our comparatively small training set, whereas low-dimensional geometric probes generalize more reliably; yet when multiple plausible corrections exist, image evidence disambiguates between solutions.

\boldparagraph{Explicit HSI losses are redundant.} Adding them on top of GRAFT yields only mixed effects: a V2S loss raises F1 on RICH ($0.565{\rightarrow}0.597$) but lowers it on PROX ($0.572{\rightarrow}0.563$), and a contact+penetration loss reverses this. This supports our design (Sec.~\ref{sec:training_objectives}): the geometric probes already encode offsets and normals, so contact emerges implicitly from GT-pose supervision.

\section{Conclusion}
\label{sec:conclusion}

We presented GRAFT, a learned HSI prior that amortizes costly geometry-based optimization into a feed-forward transformer.
By encoding the interaction state into compact, scene-grounded tokens and refining them recurrently, GRAFT matches optimization-based interaction quality at ${\sim}100{\times}$ lower runtime while substantially outperforming existing feed-forward baselines.
Because it reasons over explicit 3D geometry, GRAFT also serves as a universal plug-in prior: in geometry-only mode atop Human3R it improves contact F1 by up to 44\% without retraining or access to image features.
Current limitations include dependence on upstream scene reconstruction and human mesh estimation quality, the rigid-scene assumption of the nearest-point geometric probes (\eg, deformable surfaces are not modeled), and difficulty with heavily occluded multi-person interactions. Finally, we train on Gaussian-perturbed states; more informative refinement trajectories could better handle complex, multi-contact interactions.

\boldparagraph{\emph{Acknowledgements.}} 
We thank the whole RVH team for the support, and especially Chuqiao Li for creating the GRAFT logo. 
The authors thank the International Max Planck Research School for Intelligent Systems (IMPRS-IS) for supporting PYM and YX.
YC is funded by the Westlake Education Foundation.
NK was supported by Bosch Industry on Campus Lab at the University of Tübingen. NK thanks the European Laboratory for Learning and Intelligent Systems (ELLIS) PhD program for support.
IS and GPM were supported by the German Federal Ministry of Education and Research (BMBF): Tübingen AI Center, FKZ: 01IS18039A, by the Deutsche Forschungsgemeinschaft (DFG, German Research Foundation) -- 409792180 (Emmy Noether Programme, project: Real Virtual Humans).
GPM is a member of the Machine Learning Cluster of Excellence, EXC number 2064/1 -- Project number 390727645 and is supported by the Carl Zeiss Foundation.

\paragraph{Author contributions.}
PYM led the project, including the core idea development, method design, implementation, experimental study, and writing. YX provided close supervision and contributed to project and method design, and writing. IS provided supervision and contributed to method design and writing. YC provided valuable insights for debugging Human3R baselines and helped with writing. NK supported the use of InHabit in this work. GPM supervised the project and contributed to idea development, project framing, and writing.

\ifincludesupplementary
\providecommand{\supplementarypagebreak}{\clearpage}
\providecommand{\supplementarytitle}{\section*{Supplementary Material}}
\providecommand{\supplementarypagebreak}{}
\supplementarypagebreak
\appendix
\setcounter{section}{0}
\setcounter{figure}{0}
\setcounter{table}{0}
\setcounter{equation}{0}
\setcounter{algorithm}{0}
\renewcommand{\thesection}{S\arabic{section}}
\renewcommand{\thefigure}{S\arabic{figure}}
\renewcommand{\thetable}{S\arabic{table}}
\renewcommand{\theequation}{S\arabic{equation}}
\renewcommand{\thealgorithm}{S\arabic{algorithm}}
\makeatletter
\@ifundefined{theHsection}{}{\renewcommand{\theHsection}{supp.\arabic{section}}}
\@ifundefined{theHfigure}{}{\renewcommand{\theHfigure}{supp.fig.\arabic{figure}}}
\@ifundefined{theHtable}{}{\renewcommand{\theHtable}{supp.tab.\arabic{table}}}
\@ifundefined{theHequation}{}{\renewcommand{\theHequation}{supp.eq.\arabic{equation}}}
\@ifundefined{theHalgorithm}{}{\renewcommand{\theHalgorithm}{supp.alg.\arabic{algorithm}}}
\@ifundefined{theHALG@line}{}{\renewcommand{\theHALG@line}{supp.algline.\thealgorithm.\arabic{ALG@line}}}
\makeatother
\providecommand{\supplementarytitle}{}
\supplementarytitle

\begin{figure}[ht]
    \centering
    \vspace{-0.8em}
    \includegraphics[width=\textwidth]{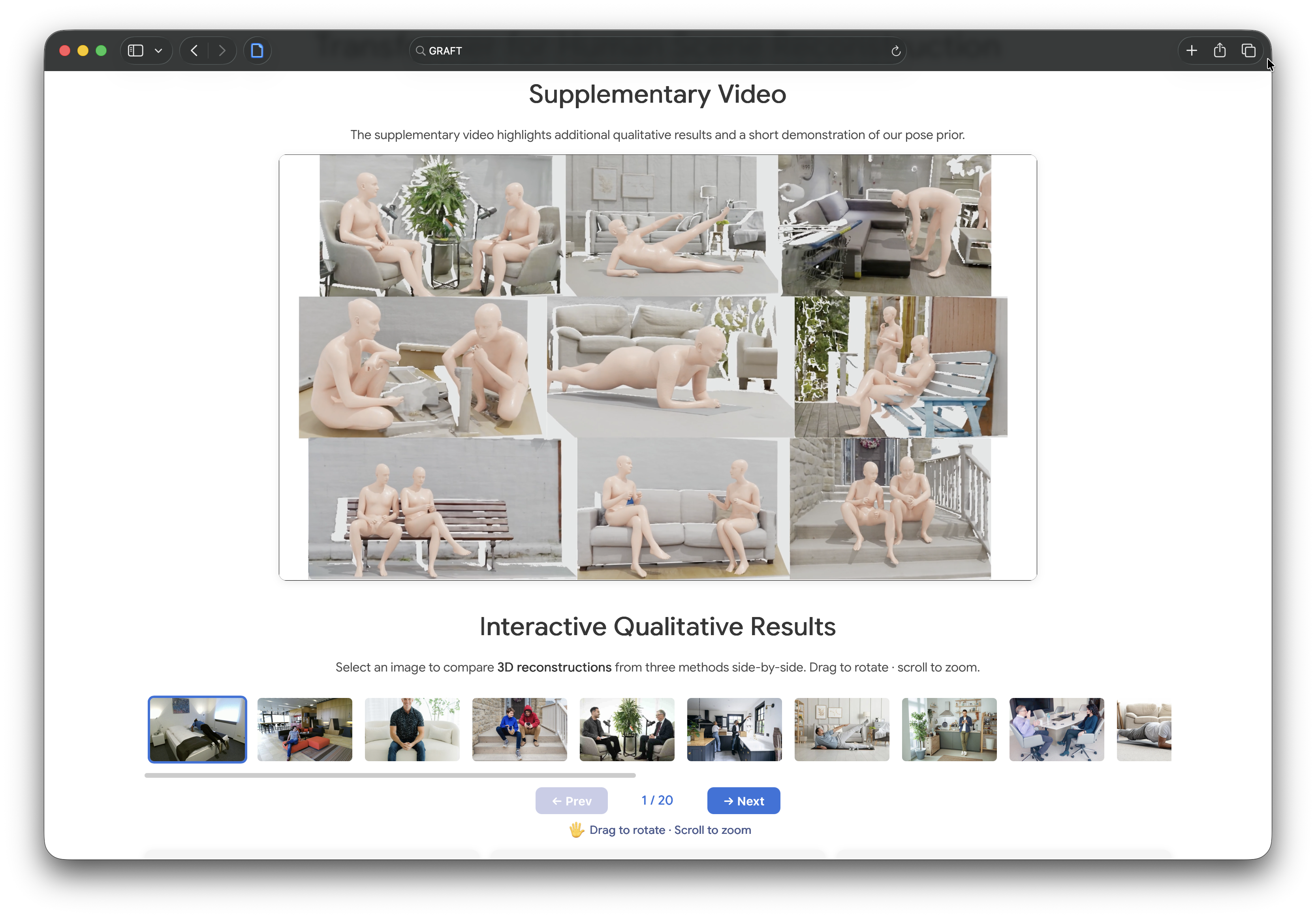}
    \vspace{-0.5em}
    \caption{\textbf{Supplementary webpage}. The webpage includes the supplementary video with reconstruction renderings, a demonstration of our pose prior, and interactive 3D results and method comparisons.}
    \label{fig:suppl_website}
    \vspace{-0.8em}
\end{figure}

We encourage readers to visit the supplementary webpage shown in Fig.~\ref{fig:suppl_website}, which presents the supplementary video with reconstruction renderings, a demonstration of our pose prior, and interactive 3D results and comparisons.
This supplementary material provides additional details complementing the main paper.
Sec.~\ref{sec:definition} gives the etymology of the GRAFT acronym.
Sec.~\ref{sec:qual} provides additional qualitative comparisons.
Sec.~\ref{sec:robustness} analyzes robustness to noisy initialization.
Sec.~\ref{sec:user_study_details} describes the user study setup and interface.
Sec.~\ref{sec:algorithm} presents the full inference and training algorithms.
Sec.~\ref{sec:scale} derives the fast differentiable scale update used during rollout training.
Sec.~\ref{sec:metrics} defines the V2S and D2S interaction metrics.
Sec.~\ref{sec:impl} details hyperparameters, loss weights, and the model architecture.
Sec.~\ref{sec:broader_context} situates GRAFT within adjacent 3D human and scene reconstruction settings.

\section{GRAFT: Definition}
\label{sec:definition}

\textbf{graft} \textit{verb}. \\
1. To join or integrate an element with another so as to bring about a close union. \\[0.5em]

\noindent\emph{Here:} aligning human pose with scene geometry for physically plausible interaction. The name reflects the core idea: the model iteratively refines human--scene interactions by predicting corrective updates to pose, translation, and shape, yielding reconstructions that are better grounded in geometry.

\section{Additional Qualitative Results}
\label{sec:qual}

Fig.~\ref{fig:suppl_qual_results} provides additional side-by-side comparisons of GRAFT, Human3R, and UniSH on in-the-wild scenes. Baselines frequently fail to capture fine interaction structure near contact regions, leading to hover artefacts or penetration, while GRAFT's iterative geometric refinement yields more accurate interactions.

\begin{figure*}[p]
    \centering
    \includegraphics[width=0.995\textwidth]{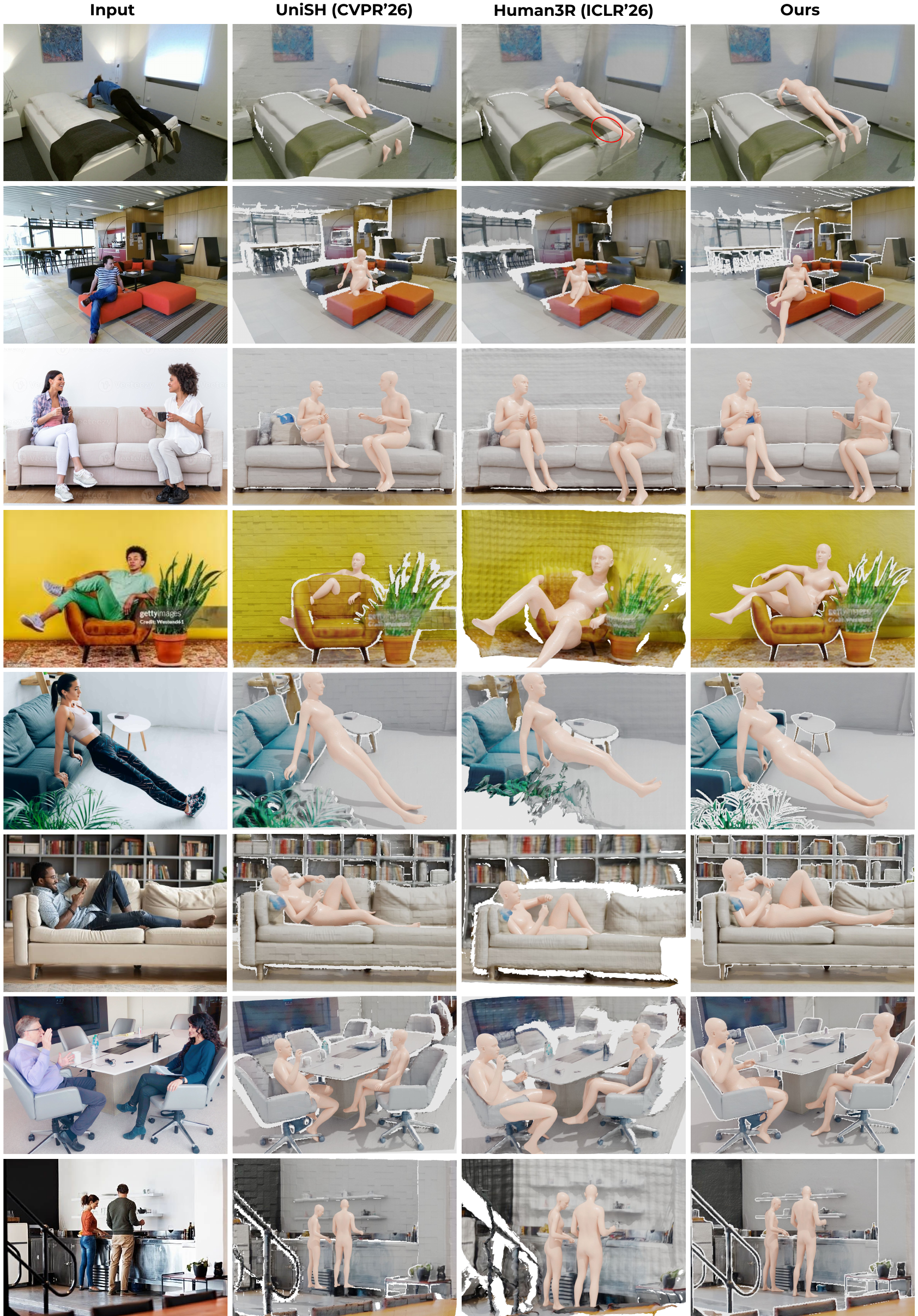}
    \caption{Qualitative comparison with Human3R and UniSH. Each row shows the scene image, followed by the reconstructions produced by each method. GRAFT (right) achieves tighter foot--ground contact and more accurate global placement, particularly in cluttered or geometrically complex environments.}
    \label{fig:suppl_qual_results}
\end{figure*}

\section{Robustness to Initialization}
\label{sec:robustness}

GRAFT's coarse alignment (Sec.~\ref{sec:HSI_initialization}) removes large-scale misalignment, leaving local adjustments to the refinement steps; any remaining gross error originates from the frozen depth backbone, which affects prior feed-forward methods equally. To probe robustness directly, we corrupt the metric initialization with Gaussian depth shifts of increasing standard deviation $\sigma$ and re-evaluate on PROX-test.

Fig.~\ref{fig:sensitivity} reports MPJPE and contact F1 as a function of $\sigma$ for both the perturbed initialization (\emph{init}) and GRAFT's refined output (\emph{Ours}). GRAFT recovers a large fraction of the injected error at every noise level, reducing MPJPE by roughly $50$--$60\%$ relative to the corrupted initialization. Contact F1 is especially stable: at $\sigma{=}1$\,m it drops only marginally ($0.56 \rightarrow 0.54$), and it remains around $0.34$ even at an extreme $\sigma{=}5$\,m, still exceeding the noise-free F1 of UniSH ($0.236$) and Human3R ($0.268$). This indicates that GRAFT's interaction quality is driven by the geometry-grounded refinement loop rather than by the initialization.

\begin{figure}[t]
    \centering
    \includegraphics[width=\linewidth]{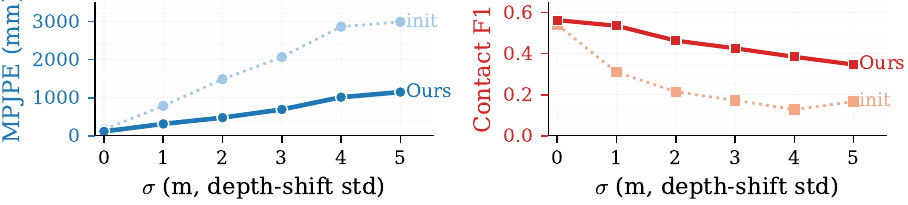}
    \caption{\textbf{Robustness to initialization noise (PROX-test).} We perturb the metric initialization with Gaussian depth shifts of standard deviation $\sigma$. As $\sigma$ grows, the initialization (\emph{init}) degrades sharply, while GRAFT (\emph{Ours}) recovers much of the error: MPJPE (left) stays far lower, and contact F1 (right) remains high, exceeding the noise-free F1 of feed-forward baselines even at $\sigma{=}5$\,m.}
    \label{fig:sensitivity}
\end{figure}

\section{User Study Details}
\label{sec:user_study_details}

Participants ($n{=}53$) were each shown 20 scenes randomly sampled from a pool of PROX-D, PiGraphs, and ${\sim}100$ in-the-wild internet images, giving each participant a unique session.
For each scene, all three reconstructions (GRAFT, UniSH, Human3R) were displayed side by side as interactive 3D viewers in a browser, with method order randomised and model names hidden.
Participants selected the reconstruction with the most plausible human--scene interaction based on contact realism and pose accuracy; see Fig.~\ref{fig:study_screenshot} for the interface.

\begin{figure}[ht]
    \centering
    \includegraphics[width=\linewidth]{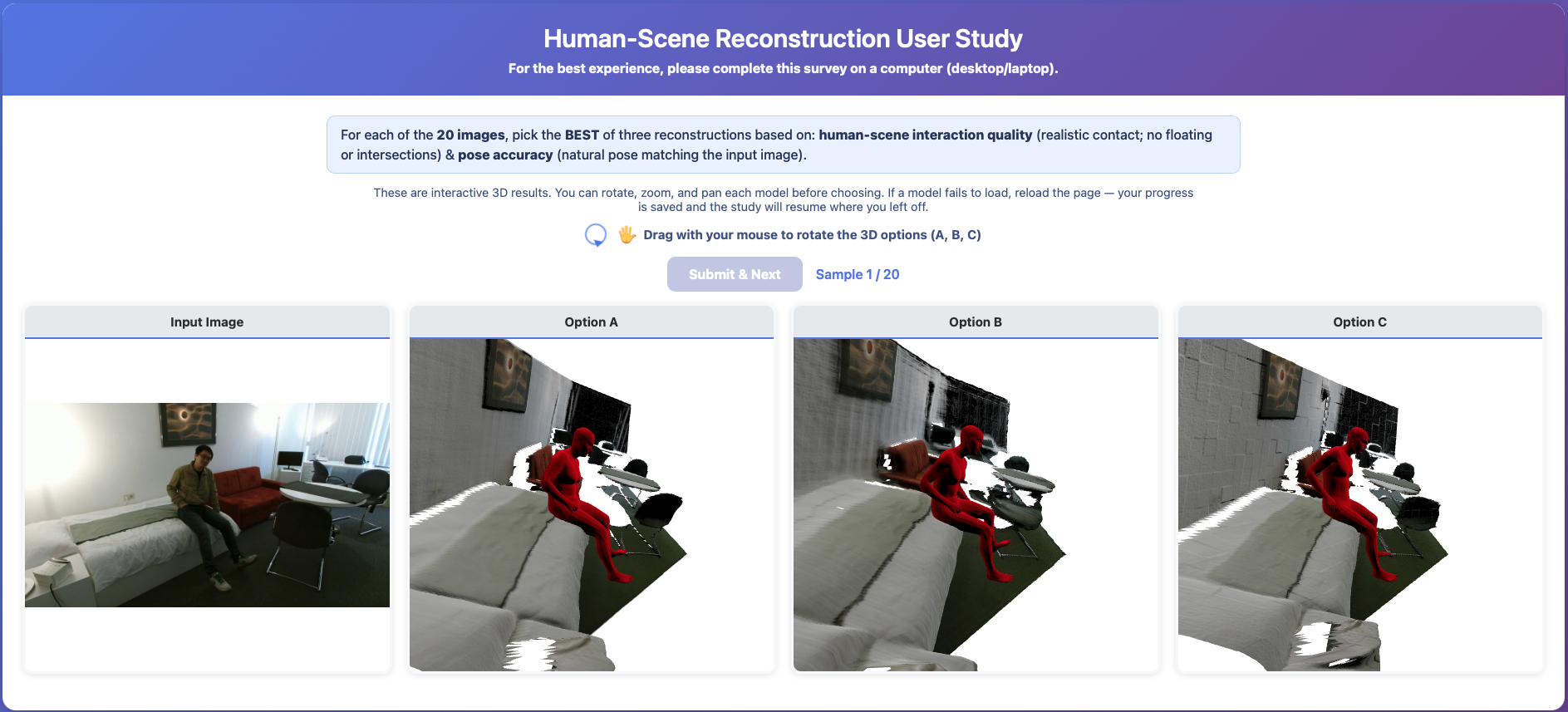}
    \caption{User study interface. The input image (left) is shown alongside three interactive 3D reconstructions that participants freely rotate before selecting the most plausible human--scene interaction.}
    \label{fig:study_screenshot}
\end{figure}

\section{Algorithm}
\label{sec:algorithm}

Alg.~\ref{alg:graft-inference} gives the full GRAFT inference procedure and Alg.~\ref{alg:graft-training} the training loop.

\begin{algorithm}[t]
\caption{GRAFT inference}
\label{alg:graft-inference}
\begin{algorithmic}
\Require Scene image $\set{I}_s$, human image $\set{I}_h$, refinement steps $T$
\Ensure Scene pointmap $\scene_s$, refined human parameters $\{\hparams_i^{T}\}_{i=1}^{N_h}$
\State $(\scene_s, \scene_h, \mathcal{F}_s, \mathcal{F}_h) \gets \texttt{MapAnything}(\set{I}_s, \set{I}_h)$
\State $\{\hparams_i^{\mathrm{init}}\}_{i=1}^{N_h} \gets \texttt{NLF}(\set{I}_h)$
\For{$i=1$ to $N_h$}
    \State $\mathbf{p}_{i}^{\mathrm{scene}} \gets \scene_h\bigl[\pi(\mathbf{p}_{i}^{\mathrm{head}})\bigr]$ \hfill\Comment{3D scene point at projected head location}
    \State $s_i^{0} \gets p_{i,z}^{\mathrm{scene}} / p_{i,z}^{\mathrm{head}}$
    \State $\hparams_i^{0} \gets \texttt{MetricAlign}(\hparams_i^{\mathrm{init}}, s_i^{0})$
\EndFor
\For{$t = 0$ to $T-1$}
    \For{$i=1$ to $N_h$}
        \State $\mathbf{g}_{i,k}^{t} \gets \texttt{GeometricProbe}_k(\scene_s, \hparams_i^{t}), \quad k=1,\dots,24$
        \State $\mathbf{z}_{i,k}^{t,0} \gets \phi_k([\mathbf{g}_{i,k}^{t}; \mathbf{p}_{i,k}^{t}]), \quad \mathbf{Z}_i^{t,0}=\{\mathbf{z}_{i,k}^{t,0}\}_{k=1}^{24}$
        \State $\mathbf{V}_i^{t} \gets \{\mathbf{v}_{i,k}^{t}\}_{k=1}^{24} \gets \texttt{VisualAnchors}(\mathcal{F}_s, \mathcal{F}_h, \hparams_i^{t})$
        \For{$\ell=1$ to $5$}
            \State $\widetilde{\mathbf{Z}}_i^{t,\ell} \gets \texttt{SelfAttn}_{\ell}(\mathbf{Z}_i^{t,\ell-1})$
            \State $\mathbf{Z}_i^{t,\ell} \gets \texttt{GeomCrossAttn}_{\ell}(\widetilde{\mathbf{Z}}_i^{t,\ell}, \mathbf{V}_i^{t})$
        \EndFor
        \State $(\dhparams_i^{t}, s_i^{t}) \gets \Psi(\mathbf{Z}_i^{t,5})$ \hfill\Comment{interaction gradient}
        \State $\hparams_i^{t+1} \gets \hparams_i^{t} + \dhparams_i^{t}$
        \State $\mathcal{M}_i^{t+1} \gets s_i^{t} \, \operatorname{SMPLX}(\hparams_i^{t+1})$
    \EndFor
\EndFor
\State \Return $\scene_s$ and $\{\hparams_i^{T}\}_{i=1}^{N_h}$
\end{algorithmic}
\end{algorithm}

\begin{algorithm}[t]
\caption{GRAFT training}
\label{alg:graft-training}
\begin{algorithmic}
\Require Batch $\mathcal{B}$, iteration $k$, rollout horizon $T{=}3$
\Ensure Updated GRAFT weights
\State $(\set{I}_s, \set{I}_h, \scene_s, \mathcal{F}_s, \mathcal{F}_h) \gets \texttt{PrepareBatch}(\mathcal{B})$
\State $(\hparams^{*}, \mathcal{V}^{*}, \widetilde{\mathcal{V}}^{*}) \gets \texttt{PrepareGT}(\mathcal{B})$
\State $\hparams^{\mathrm{nlf}} \gets \texttt{NLFInit}(\mathcal{B})$
\State Sample $\hparams^{0}$ from $\{\hparams^{\mathrm{nlf}},\; \hparams^{*},\; \hparams^{*} + \epsilon\}$, $\;\epsilon \sim \mathcal{N}(0, \Sigma)$ \hfill\Comment{NLF / GT / perturbed GT}
\State Apply visual-anchor dropout to sampled anchor neighbourhoods
\State $R \gets 1$ if $k < 10\mathrm{k}$, else $R \gets T$
\State $\mathcal{L} \gets 0$
\For{$t = 0$ to $R-1$}
    \State $(\dhparams^{t}, s^{t}) \gets \modelName_{\omega}(\scene_s, \mathcal{F}_s, \mathcal{F}_h, \hparams^{t})$ \hfill\Comment{shared weights over $t$}
    \State $\hparams^{t+1} \gets \hparams^{t} + \dhparams^{t}$
    \State $\mathcal{V}^{t+1} \gets s^{t} \, \operatorname{SMPLX}(\hparams^{t+1})$
    \State $\widetilde{\mathcal{V}}^{t+1} \gets \mathcal{V}^{t+1} - \operatorname{mean}(\mathcal{V}^{t+1})$
    \State $\mathcal{L} \gets \mathcal{L} + \lambda_p\lVert \pose^{t+1} - \pose^{*} \rVert_2^2 + \lambda_v\lVert \mathcal{V}^{t+1} - \mathcal{V}^{*} \rVert_2^2 + \lambda_n\lVert \widetilde{\mathcal{V}}^{t+1} - \widetilde{\mathcal{V}}^{*} \rVert_2^2$
\EndFor
\State $\omega \gets \texttt{Adam}(\omega, \nabla_{\omega} \mathcal{L})$
\State Update learning rate by linear warmup followed by cosine decay
 \end{algorithmic}
\end{algorithm}

\section{Fast Differentiable Scale Update}
\label{sec:scale}

\boldparagraph{Why an explicit scale.} Human--scene mismatches are dominated by scale ambiguity: monocular 2D alignment is typically accurate, so the residual error after initialization is largely metric. We therefore expose a single uniform scale $s$ as an explicit output, decoupled from body proportions ($\shape$) and translation ($\trans$). Correcting scale through shape alone would instead couple the update across all 10 shape coefficients (and $\trans$) at every step. As illustrated in Fig.~\ref{fig:scale_refit}, the predicted $s_t$ is folded back into the shape coefficients via the closed-form update derived below and fed to the next refinement iteration.

\begin{figure}[t]
    \centering
    \includegraphics[width=\linewidth]{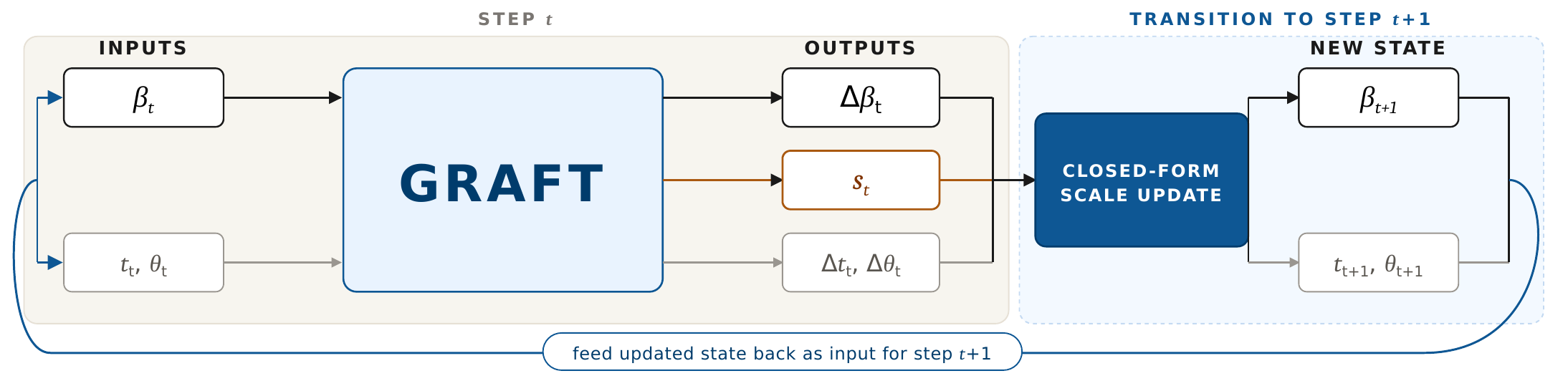}
    \caption{\textbf{Fast differentiable scale update.} At each step, GRAFT predicts pose and translation updates ($\Delta\pose_t, \Delta\trans_t$) together with a uniform scale $s_t$. Rather than re-fitting SMPL-X to explicitly scaled vertices, our closed-form update folds $s_t$ directly into the shape coefficients ($\shape_t \rightarrow \shape_{t+1}$); the rescaled body is recovered by simple vector arithmetic and fed back as the input state for step $t{+}1$.}
    \label{fig:scale_refit}
\end{figure}

\boldparagraph{Derivation.} Since uniform scaling leaves pose rotations unchanged, we ignore pose and define a simplified SMPL-X parameterization:
\begin{equation}
\label{eq:smplx-simplified}
    \operatorname{SMPLX}(\shape) = \mathbf{T} + \mathbf{S}^\top\shape,
\end{equation}
where $\mathbf{T} \in \mathbb{R}^{N}$ is the mean body template in a canonical pose ($N = 10475 \times 3$), $\shape \in \mathbb{R}^{10}$ are the shape coefficients, and $\mathbf{S} \in \mathbb{R}^{10 \times N}$ are the shape blend shape directions. Given a predicted scale factor $s$, we seek scaled shape coefficients $\shape_s$ satisfying:
\begin{equation}
\label{eq:scaling-condition}
    \mathbf{T} + \mathbf{S}^\top\shape_s = s\bigl(\mathbf{T} + \mathbf{S}^\top\shape\bigr).
\end{equation}
Were $\mathbf{T}$ absent, the solution would be trivially $\shape_s = s\,\shape$. The template introduces an additive offset that must be absorbed into $\shape$-space.

To this end, we project $\mathbf{T}$ into shape space via least squares:
\begin{equation}
\label{eq:lsq}
    \mathbf{c} = \arg\min_{\mathbf{c}} \| \mathbf{S}^\top\mathbf{c} - \mathbf{T} \|_2^2 = (\mathbf{S}\mathbf{S}^\top)^{-1}\mathbf{S}\mathbf{T},
\end{equation}
so that $\mathbf{T} \approx \mathbf{S}^\top\mathbf{c}$. Substituting this approximation into Eq.~\eqref{eq:scaling-condition}:
\begin{equation}
\label{eq:substituted}
    \mathbf{S}^\top\mathbf{c} + \mathbf{S}^\top\shape_s = s\bigl(\mathbf{S}^\top\mathbf{c} + \mathbf{S}^\top\shape\bigr).
\end{equation}
Factoring out $\mathbf{S}^\top$ on both sides and cancelling it (as $\mathbf{S}$ has full row rank), we obtain:
\begin{equation}
\label{eq:scale-update}
    \shape_s = s \cdot (\shape + \mathbf{c}) - \mathbf{c}.
\end{equation}
This is a closed-form, fully differentiable update: $\mathbf{c}$ is computed offline once per body model, translations scale directly, and rotations are unchanged. It replaces expensive per-step shape refitting with simple vector arithmetic, enabling efficient multi-step rollout supervision.

\section{Interaction Metric Details: V2S and D2S}
\label{sec:metrics}

We introduce two geometry-continuous metrics to complement the hard-threshold contact F1 score.
Both are computed over the PROX contact-vertex set: approximately 1000 vertices distributed across body regions commonly involved in human--scene contact, including the hands, feet, back, and thighs.

\boldparagraph{Vector-to-Scene (V2S).}
For each contact vertex $v$ in the set, we compute the 3D displacement vector $\mathbf{d}(v) = \mathbf{p}^*_v - \mathbf{p}_v$ from the vertex position to its nearest scene point, both in the prediction and in the ground truth.
V2S reports the weighted mean Euclidean error between these predicted and ground-truth displacement vectors (in mm):
\begin{equation}
\text{V2S} = \frac{\sum_v w_v \, \lVert \mathbf{d}^{\text{pred}}(v) - \mathbf{d}^{\text{gt}}(v) \rVert_2}{\sum_v w_v}.
\end{equation}

\boldparagraph{Direction-to-Scene (D2S).}
D2S reports the weighted mean angular error between the predicted and ground-truth displacement vectors (in degrees):
\begin{equation}
\text{D2S} = \frac{\sum_v w_v \, \angle\!\left(\mathbf{d}^{\text{pred}}(v),\, \mathbf{d}^{\text{gt}}(v)\right)}{\sum_v w_v}.
\end{equation}
Together, V2S captures errors in the magnitude and direction of human--scene proximity, while D2S isolates directional misalignment independently of scale.

\boldparagraph{Density-aware vertex weighting.}
The SMPL-X mesh is non-uniform: hand regions contain far more vertices than coarser regions such as the thigh.
Without correction, high-density regions would dominate both metrics regardless of physical significance.
Each vertex $v$ is therefore weighted by its Voronoi area on the mesh surface, so that dense regions (e.g. fingers) contribute less and coarser regions (e.g. thigh) contribute more, making the metrics independent of tessellation density.

\section{Implementation Details}
\label{sec:impl}

\boldparagraph{Learning-rate schedule.} We train with a batch size of $16$. Linear warmup runs for the first $2$k iterations (from $0$ to $1\times10^{-4}$), followed by cosine decay to $2\times10^{-7}$ over the remaining $213$k iterations.

\boldparagraph{Loss weights.} Camera-relative vertex loss weight $\lambda_v=7.0$; centered-vertex loss weight $\lambda_n=5.0$. Rotation weights per parameter group: global orientation $5.0$, body pose $2.0$, left-hand pose $0.5$, right-hand pose $0.5$. Translation and body shape are supervised implicitly through the vertex losses. No explicit contact or interpenetration losses are used.

\boldparagraph{Geometric feature encoding.}
Each geometric probe yields three 3D quantities: the displacement to the nearest scene point, the surface normal at that point, and the probe position in body-relative coordinates.
The surface normal at a scene point is computed as the mean of the cross-products of its four pointmap neighbours, taken anticlockwise and oriented to face the camera.
Each is encoded by a separate learnable Fourier feature encoder ($64$ output dimensions, with the raw input concatenated), and the results are concatenated with the current joint state before being lifted to the transformer dimension by a two-layer MLP.
Hand and full-body tokens additionally compress their per-probe features through a shared small MLP before concatenation.
We also experimented with assigning a separate token to each probe instead of grouping probes into body-joint, hand, and full-body tokens; this enlarged the token count without improving accuracy, so we retain the compact 24-token layout.

\boldparagraph{Model architecture.}
GRAFT uses a transformer width of $512$ with $5$ layers and $8$ attention heads, totalling $16.2$M parameters across transformer layers, tokenizer MLPs, and decoder heads---a deliberately lightweight design.
Visual features are sampled from four levels of MapAnything: the post-ViT output, two intermediate activations within the alternating transformer, and its final output.
At each spatial position in the sampled neighbourhood, the four level features are each linearly projected to $128$ dimensions, concatenated, and passed through a $512{\to}512$ linear layer, yielding one $512$-dimensional token per position.
For body and hand anchors we sample a $3{\times}3$ neighbourhood, producing $9$ tokens per stream; for the 27 full-body surface anchors we sample a single token each ($1{\times}1$).
Features are extracted from two streams---the scene image $\set{I}_s$ and the interaction image $\set{I}_h$---so each HSI token cross-attends to $2n^2$ context tokens in total.
A learnable \emph{stream embedding} (one per stream, added after the per-level fusion) lets the model distinguish scene from interaction evidence.
All decoder heads $\Psi$ are two-layer MLPs with hidden dimension $256$, and all MLPs throughout the model use GELU activations.

\section{Broader Context}
\label{sec:broader_context}

GRAFT's geometry-grounded refinement connects to a broader effort on modelling 3D humans under partial or indirect observation. One active direction generates or synthesizes 3D humans, whether from large synthetic corpora~\cite{black2023bedlam, tesch2025bedlam2}, conditioned on 3D scenes~\cite{wang2022humanise, zhang2024scenic, jiang2024truman}, or reconstruct from sensor observation~\cite{ym2025physic, chen2025human3r, xue2025gen3diffusion, xue2025infinihuman, xue2022event, xue2024elnr}; such generative priors are complementary to GRAFT's corrective, scene-grounded refinement.

\fi

\bibliographystyle{splncs04}
\bibliography{main}
\end{document}